\documentclass[utf8]{article} 
\usepackage{pgfplots}
\pgfplotsset{compat=1.8}
\usetikzlibrary{shapes,arrows}
\usepackage{url,hyperref,subcaption}
\usepackage[onehalfspacing]{setspace}

\usepackage{inputenc}
\usepackage{graphicx}
\usepackage[margin=0.5in]{geometry}
\usepackage{amsmath}
\usepackage{amsthm}
\usepackage{amsfonts}
\usepackage{amssymb}
\usepackage{array}
\usepackage{algorithm}
\usepackage{algorithmic}
\usepackage{mathrsfs}
\usepackage{natbib}
\setcitestyle{authoryear,open={(},close={)}} 
\usepackage{enumerate}
\usepackage{lipsum}
\usepackage{booktabs} 
\usepackage{dsfont}
\usepackage{latexsym}
\usepackage{hhline}
\usepackage{color}
\usepackage{textcomp}
\usepackage[normalem]{ulem}

\newtheorem{lemma}{Lemma}
\newtheorem{definition}{Definition}

\newtheorem{proposition}{Proposition}

\makeatletter
\newcommand{\argmin}{\mathop{\text{~argmin~}}}

\DeclareMathOperator{\sparkn}{spark}
\DeclareMathOperator{\Supp}{Supp}

\DeclareMathOperator{\prox}{prox}
\DeclareMathOperator{\vecn}{vec}

\definecolor{gris}{gray}{0.90}
\definecolor{gris25}{gray}{0.90}
\definecolor{americanrose}{rgb}{1.0, 0.01, 0.24}
\definecolor{bostonuniversityred}{rgb}{0.8, 0.0, 0.0}
\definecolor{shamrockgreen}{rgb}{0.0, 0.62, 0.38}
\definecolor{selectiveyellow}{rgb}{1.0, 0.73, 0.0}
\definecolor{royalblue}{rgb}{0.25, 0.41, 0.88}
\definecolor{ashgrey}{rgb}{0.7, 0.75, 0.71}
\definecolor{burgundy}{RGB}{159,29,53}
\definecolor{darkgreen}{RGB}{18,53,26}
\definecolor{lightblue}{RGB}{102,217,255}
\definecolor{fakeorange}{RGB}{255,140,102}
\definecolor{arylideyellow}{rgb}{0.91, 0.84, 0.42}
\definecolor{bananayellow}{rgb}{1.0, 0.88, 0.21}
\definecolor{gris_f}{gray}{0.35}
\definecolor{bordure}{rgb}{0.09,0.17,0.68}
\definecolor{aquamarine}{rgb}{0.5, 1.0, 0.83}
\definecolor{apricot}{rgb}{0.98, 0.81, 0.69}
\definecolor{babyblue}{rgb}{0.54, 0.81, 0.94}
\definecolor{uipoppy}{RGB}{225, 64, 5}
\definecolor{uipaleblue}{RGB}{96,123,139}
\definecolor{uiblack}{RGB}{0, 0, 0}
\definecolor{decoda}{RGB}{0,153, 0}
\definecolor{lightgreen}{rgb}{0.56, 0.93, 0.56}
\definecolor{blue_f}{rgb}{0.2, 0.2, 0.6}
\definecolor{cinnamon}{rgb}{0.82, 0.41, 0.12}
\definecolor{darkpastelgreen}{rgb}{0.2, 0.75, 0.24}
\definecolor{drab}{rgb}{0.59, 0.44, 0.09}

\title{Dictionary-based Low-Rank Approximations and the Mixed Sparse Coding problem}
\author{Jeremy E. Cohen\\
        Univ Lyon, INSA-Lyon, UCBL, UJM-Saint Etienne, CNRS, Inserm,\\ CREATIS UMR 5220, U1206, F-69100 Villeurbanne, France}
\date{}

\begin{document}

\maketitle

\begin{abstract}
  Constrained tensor and matrix factorization models allow to extract interpretable patterns from multiway data. Therefore
  crafting efficient algorithms for constrained low-rank approximations is nowadays an
  important research topic.
  This work deals with columns of factor matrices of a low-rank approximation being sparse in a known and possibly overcomplete basis, a model coined as Dictionary-based Low-Rank Approximation (DLRA).
  While earlier contributions focused on finding factor columns inside a dictionary of candidate columns, \textit{i.e.} one-sparse approximations, this work is the first to tackle DLRA with sparsity larger than one. I propose to focus on the sparse-coding subproblem coined Mixed Sparse-Coding (MSC) that emerges when solving DLRA with an alternating optimization strategy. Several algorithms based on sparse-coding heuristics (greedy methods, convex relaxations) are provided to solve MSC. The performance of these heuristics is evaluated on simulated data. Then, I show how to adapt an efficient MSC solver based on the LASSO to compute Dictionary-based Matrix Factorization and Canonical Polyadic Decomposition in the context of hyperspectral image processing and chemometrics. These experiments suggest that DLRA extends the modeling capabilities of low-rank approximations, helps reducing estimation variance and enhances the identifiability and interpretability of estimated factors.

\end{abstract}

\section{Introduction}

Low-Rank Approximations (LRA) are well-known dimensionality reduction techniques that allow to represent tensors or matrices as sums of a few separable terms. One of the main reasons why these methods are used extensively for pattern recognition is their ability to provide part-based representations of the input data. This is particularly true for Nonnegative Matrix Factorization (NMF) or Canonical Polyadic Decomposition (CPD), see Section~\ref{sec:background} for a quick introduction and the following surveys and book for more details~\citep{Kolda2009Tensor, sidiropoulos2017tensor, gillis2020nonnegative}. In order to fix notations while retaining generality, let us make use of the following informal mathematical formulation of LRA.

\begin{definition}[Low-Rank Approximation]
   Given a data matrix \( Y\in\mathbb{R}^{n\times m} \) and a small nonzero rank \( r\in\mathbb{N} \), compute so-called factor matrices \( A\in\mathbb{R}^{n\times r} \) and \( B\in\mathbb{R}^{m\times r} \) in
\begin{equation}\label{eq:LRA}
  \underset{A\in\Omega_A,\;\;B\in\Omega_B}{\argmin} \| Y-AB^T \|_F^2~.
\end{equation}
The set \( \Omega_A \) is an additional constraint set for \( A \), such as nonnegativity elementwise. The set \( \Omega_B \) is the structure required on \( B \) to obtain a specific LRA model; for instance in the unconstrained CPD of an order three tensor, \( \Omega_B\) is the set of matrices written as the Khatri-Rao product of two factor matrices. A precise qualification of ``small'' rank depends on the sets \( \Omega \) and is omitted for simplicity. 
\end{definition}

 Among LRA models, identifiability\footnote{Informally, the parameters of a model are identifiable if they can be uniquely recovered from the data.} properties vary significantly. While CPD is usually considered to have mild identifiability conditions~\citep{Kruskal1977Three}, NMF on the other hand is often not essentially unique~\citep{gillis2020nonnegative}\footnote{essential uniqueness means uniqueness up to trivial scaling ambiguities and rank-one terms permutation.}. Generally speaking, additional regularizations are often imposed in both matrix and tensor LRA models to help with the interpretability of the part-based representation. These regularizations may take the form of constraints on the parameters or penalizations, like the sparsity-inducing \( \ell_1 \) norm~\citep{Hoyer2002Non, morup2008algorithms}. They can also take the form of
 a parameterization of the factors~\citep{hennequin2010time, Timmerman2002Three} if such a parameterization is known.

This work focuses on imposing an implicit parameterization on \( A \). More precisely, each column of factor \( A \) is assumed to be well represented in a known dictionary \( D\in\mathbb{R}^{n\times d} \) using only a small number \( k < n \) of coefficients. In other words, in this work the set \( \Omega_A \) is the union of subspaces of dimension \( k \) spanned by columns of a given dictionary \( D \), and there exist a columnwise k-sparse code matrix \( X\in\mathbb{R}^{d\times r} \) such that \( A = DX \). A low-rank approximation model which such a constraint on at least one mode is called Dictionary-based LRA (DLRA) thereafter.

\begin{definition}[Dictionary-based Low-Rank Approximation]
     Given a data matrix \( Y\in\mathbb{R}^{n\times m} \), a
      small nonzero rank \( r\in\mathbb{N} \), a sparsity level \( k<n \) and a dictionary \( D\in\mathbb{R}^{n\times d} \), compute columnwise \(k\)-sparse code matrix \( X\in\mathbb{R}^{n\times r} \) and factor matrix \( B\in\mathbb{R}^{m\times r} \) in
  \begin{equation}\label{eq:DLRA}
    \underset{X\in\Omega_X,\;\;\forall i\leq r, \; \|X_i\|_0\leq k,\;\;  B\in\Omega_B}{\argmin} \| Y-DXB^T \|_F^2~,
  \end{equation}
  where \( X_i \) is the \(i\)-th column of \( X \).
  The set \( \Omega_X \) is an additional constraint set for \( X \), such as nonnegativity elementwise. The set \( \Omega_B \) is the structure required on \( B \) to obtain a specific LRA model. Additionally, assume that \( B \) is full column-rank.
\end{definition}

\subsection{Motivations}\label{sec:motivation}

To better advocate for the usefulness of DLRA, below two particular DLRA models are introduced:
\begin{itemize}
  \item \underline{Dictionary-based Nonnegative Matrix Factorization (DNMF)} may be formulated as
  \begin{equation}\label{eq:DNMF}
    \underset{X\in\mathbb{R}_+^{d\times r},\;\forall i\leq r, \; \|X_i\|_0\leq k,\;\; B\in\mathbb{R}_+^{m\times r}}{\argmin} \| Y-DXB^T \|_F^2~.
  \end{equation}
  To ensure that \(A=DX \) is nonnegative, dictionary \( D \) is supposed to be entry-wise nonnegative so that the constraint \( X\geq 0 \) is sufficient. NMF is known to have strict identifiability conditions~\citep{fu2018identifiability,gillis2020nonnegative}, and in general one should expect that NMF has infinitely many solutions. The dictionary constraint can enhance identifiability by restraining the set of solutions. For instance setting \( D=Y \) and \( k=1 \) yields the Separable NMF model, which is solvable in polynomial time and which factors are generically identifiable even in the presence of noise~\citep{Gillis2014Robust}.
  Moreover, DNMF is also a more flexible model than NMF. In Section~\ref{sec:xpDLRA}, it is shown that DNMF can be used to solve a matrix completion problem with missing rows in the data that NMF cannot solve.

  \item \underline{Dictionary-based Canonical Polyadic Decomposition (DCPD)}, for an input tensor \( Y\in\mathbb{R}^{n\times m_1\times m_2} \) may be formulated as
  \begin{equation}\label{eq:DCPD}
    \underset{X\in\mathbb{R}^{d\times r},\;\forall i\leq r, \; \|X_i\|_0\leq k,\;\; B\in\mathbb{R}^{m_1\times r},\; C\in\mathbb{R}^{m_2\times r}}{\argmin} \| Y-DX\left(B\odot C\right)^T \|_F^2~.
    \end{equation}
  where \( \odot \) is the Khatri-Rao product~\citep{Kolda2009Tensor}. Unlike NMF, CPD factor are identifiable under mild conditions~\citep{Kruskal1977Three}. But in practice, identifiability may still be elusive, and the approximation problem is in general ill-posed~\citep{Silva2008Tensor} and algorithmically challenging~\citep{mohlenkamp2019dynamics}. Therefore the dictionary constraint can be used to reduce estimation variance given an adequate choice of dictionary \( D \). It was shown when \( k=1 \) in Equation~\eqref{eq:DCPD} that it enhances identifiability of the CPD and makes the optimization problem well-posed~\citep{Cohen2018Dictionary}.
\end{itemize}

Note that these models have known interesting properties when \( k=1 \), but have not been studied in the general case.
A motivation for this work is therefore to expand these previous works to the general case \(n> k\geq 1 \), focusing on the algorithmic aspects. Section~\ref{sec:background} provides more details on the one-sparse case.


\subsection{Contributions}

The first contribution of this work is to propose the new DLRA framework. The proposed DLRA allows to constrain low-rank approximation problems so that some of the factor matrices are sparse columnwise in a known basis. This includes sparse coding the patterns estimated by constrained factorization models such as NMF, or tensor decomposition models such as CPD.
The impact of the dictionary constraint on the LRA parameter estimation error is studied experimentally in Section~\ref{sec:xpDLRA} dedicated to experiments with DLRA, 
where the flexibility of DLRA is furthermore illustrated on real applications. I show that DLRA allows to complete entirely missing rows in incomplete low-rank matrices. I also show the advantage of finding the best atoms algorithmically when imposing smoothness in DCPD using B-splines.

A second contribution is to design an efficient algorithm to solve Problem~\eqref{eq:DLRA}. To that end, I shall focus on (approximate) Alternating Optimization (AO), understood as Block Coordinate Descent (BCD) where each block update consists in minimizing almost exactly the cost with respect to the updated block while other parameters are fixed. There are two reasons for this choice. First, AO algorithms are known to perform very well in many LRA problems. They are state-of-the-art for NMF~\citep{gillis2020nonnegative} and standard for Dictionary Learning~\citep{Engan1999MethodOO, Mairal2010Online} and tensor factorization problems~\citep{Huang2016flexible, fu2020computing}.
Nevertheless inexact BCD methods and all-at-once methods are also competitive~\citep{Aharon2005K, Xu2015Alternating, kolda2020stochastic, marmin2021joint} and an inertial Proximal Alternating Linearization Method (iPALM) for DLRA is quickly discussed in Section~\ref{sec:ipalm}. The proposed algorithm is coined AO-DLRA.

As a third contribution, I study the subproblem in Problem~\eqref{eq:DLRA} with respect to \( X \).
Developing a subroutine to solve it for known \( B \) allows not only for designing AO-DLRA, but also for post-processing a readily available estimation of \( A \). In fact a significant part of this manuscript is devoted to studying the subproblem of minimizing the cost in Problem~\eqref{eq:DLRA} with respect to \( X \), which is labeled Mixed Sparse Coding (MSC),
\begin{equation}\label{eq:MSC}\tag{MSC}
  \underset{\forall i\leq r,\;\|X_i\|_0\leq k}{\argmin} \|Y - DXB^T\|_F^2 ~.
\end{equation}
In this formulation of MSC, no further constraints are imposed on \( X \) and therefore \( \Omega_X \) is \( \mathbb{R}^{d\times r} \). While similar to a matrix sparse coding, which is obtained by setting \( B \) to the identity matrix, it will become clear in this manuscript that MSC should not be handled directly using sparse coding solvers in general. For instance, it is shown in Section~\ref{sec:ortho} that while having an orthogonal dictionary \( D \) yields a polynomial time algorithm to solve MSC when \( r=1 \) using Hard Thresholding, this does not extend when \( r>1 \). In this work, the MSC problem, which is NP-hard as a generalization of sparse coding, is studied in order to build several reasonable heuristics that may be plugged into an AO algorithm for DLRA.

\subsection{Structure}
The paper is divided in three remaining sections. The first section provides the necessary background for this manuscript. The second one is devoted to studying MSC and heuristics to solve it. The third section shows how to compute DLRA using the heuristics developed in the first part. In Section~\ref{sec:propmsc}, we study formally the MSC problem and its relationship with sparse coding. In Section~\ref{sec:ncvxheur} we study a simple heuristic based solely on sparse coding each column of \( YB^{\dagger} \). Two $\ell_0$ heuristics similar to Iterative Hard Thresholding (IHT) and Orthogonal Matching Pursuit (OMP) are also introduced, while two types of convex relaxations are studied in Section~\ref{sec:cvxheur}. Section~\ref{sec:xpMSC} is devoted to compare the practical performance of the various algorithms proposed to solve MSC and shows that a columnwise \( \ell_1 \) regularization coined Block LASSO is a reasonable heuristic for MSC. Section~\ref{sec:AODLRA} shows how Block LASSO can be used to compute various DLRA models, while Section~\ref{sec:xpDLRA} illustrates DLRA on synthetic and real-life source-separation problems.

All the proofs are deferred to the supplementary material attached to this paper, as well as
the pseudo-codes of some proposed heuristics and additional experiments.

\subsection{Notations}

Vectors are denoted by small letters \( x \), matrices by capital letters \( X \). The indexed quantity \( X_i \) refers to the $i$-th column of matrix \( X \), while \( X_{ij} \) is the \((i,j)\)-th entry in \( X \). A subset \( S \) of columns of \( X \) is denoted \( X_S \), while a submatrix with columns in \( S_i \) and rows in \( S_j \) is denoted \( X_{S_iS_j} \). The submatrix of \( X \) obtained by removing the \( i \)-th column is denoted by \( X_{-i} \).
The \(i\)-th row of matrix \( X \) is denoted by \( X_{\cdot i} \) and \( X_{\cdot I} \) is the submatrix of \( X \) with rows in set \( I\).
The \( \ell_0(x)=\|x\|_0 \)
map counts the number of nonzero elements in \( x \). The product \( \otimes \) denotes the Kronecker product, a particular instance of tensor product~\citep{Brewer1978Kronecker}. The Khatri Rao product \(\odot\) is the columnwise Kronecker product. The support of a vector or matrix \( x \), \textit{i.e.} the location of its nonzero elements, is denoted by \( \Supp(x) \). If \( S=\Supp(x) \), the location of the zero elements is denoted \( \overline{S} \). The list \( [M,N,P] \) denotes the concatenation of columns of matrices \( M, N \) and \( P \). A set \( I \) contains \( |I| \) elements.

\section{Background}\label{sec:background}

Let us review the foundations of the proposed work, matrix and tensor decompositions and sparse coding, as well as existing models closely related to the proposed DLRA.

\subsection{Matrix and tensor decompositions}

Matrix and tensor decompositions can be understood as pattern mining techniques, which extract meaningful information out of collections of input vectors in an unsupervised manner. Arguably one of the earliest form of interpretable matrix factorisation is Principal Component Analysis~\cite{pearson1901liii, Jolliffe2002Principal}, which extract a few orthogonal significant patterns out of a given matrix while performing dimensionality reduction. 

Other matrix factorization models exploit other constraints than orthogonality to mine interpretable patterns. Blind source separation models such as Independent Component Analysis historically exploited statistical independence~\cite{comon1994independent}, while NMF, which assumes all parameters are elementwise nonnegative, has received significant scrutiny over the last two decades following the seminal paper of Lee and Seung~\cite{Lee1999Learning}. Sparse models such as Sparse Component Analysis exploit sparsity on the coefficients~\cite{Gribonval2006survey}. It can be noted that while all those factorization techniques aim at providing interpretable representations, they are typically identifiable under strict conditions not necessarily met in practice~\cite{Donoho2003When,fu2018identifiability,Cohen2019Identifiability}. It should be noted that the most important underlying hypothesis in matrix factorization is the linear dependency of the input data with respect to the templates/principal components stored in matrix \(A\) following the notations of Equation~\ref{eq:LRA}.

In practice, to compute for instance NMF, one solves an optimization problem of the form 
\begin{equation}\label{eq:NMFopt}
  \underset{A\geq 0,\; B\geq 0}{\argmin}~\|Y - AB^T \|_F^2
\end{equation}
which is nonconvex with respect to \(A,B\) jointly but convex for each block. A common family of methods therefore uses alternating optimization in the spirit of Alternating Least Squares~\cite{Gillis2012Accelerated}. Other loss functions can easily be used~\cite{Lee1999Learning}. However computing matrix factorization models is often a difficult task; in fact NMF and sparse component analysis are both NP-hard problems in general and existing polynomial time approximation algorithms should be considered heuristics~\cite{Natarajan1995Sparse,Vavasis2010complexity}.

Tensor decompositions follow the same ideas of unsupervised pattern mining and linearity with respect to the representation basis, but extract information out of tensors rather than matrices. Tensors in this manuscript are considered simply as multiway arrays~\cite{Kolda2009Tensor} as is usually done in data sciences. Tensors have become an important data structure as they appear naturally in a variety of applications such as chemometrics~\cite{Bro1997PARAFAC}, neurosciences~\cite{Becker2015Brain}, remote sensing~\cite{Cohen2015Fast} or deep learning~\cite{kossaifi2020tensor}. 

At least two families of tensor decomposition models can be considered, with quite different identifiability properties and applications. A first family contains intepretable models such as the Canonical Polyadic Decomposition, often called PARAFAC~\cite{hitchcock1927expression}, or closely related models such as PARAFAC2~\cite{Harshman1972PARAFAC2}. Contrarily to constrained matrix decomposition models, the addition of at least a dimension compared to matrix factorization fixes the rotation ambiguity inherent to matrix factorization models, and therefore the CPD model is identifiable under mild conditions~\cite{Kruskal1977Three,Sidiropoulos2000Uniqueness,Domanov2013uniqueness}. Nevertheless, additional constraints are commonly imposed on the parameters of these models to refine the interpretability of the parameters, reduce estimation errors or improve the properties of the underlying optimization problem~\cite{Lim2009Nonnegative,roald2021parafac2}. A second family is composed of tensor formats, in particular the Tucker decomposition~\cite{Tucker1966Some} and a wide class of tensor networks such as tensor trains~\cite{Oseledets2011Tensor}. These models are not used in general for solving inverse problems but rather for compression or dimensionality reduction. Nevertheless, they turn into interesting pattern mining tools given adequate constraints such as nonnegativity~\cite{Phan2011Extended,Marmoret2020Uncovering}.


\subsection{Sparse Coding}

Spare Coding (SC) and other sparse approximation problems typically try to describe an input vector as a sparse linear combination of well-chosen basis vectors. A typical formulation for sparse coding an input vector \(y\in\mathbb{R}^{n}\) in a code book or dictionary \(D\in\mathbb{R}^{n\times d}\) is the following nonconvex optimization problem
\begin{equation}\label{eq:SC}\tag{SC}
  \underset{\|x\|_0\leq k}{\argmin} \|y - Dx\|_2^2,
\end{equation}
where \(k \leq n\) is the largest number of nonzero entries allowed in vector \(x\), \textit{i.e.} the size of the support of \(x\). As long as the dictionary \(D\) is not orthogonal, this problem is difficult to solve efficiently and is in fact NP-hard~\cite{Natarajan1995Sparse}. The body of literature of algorithms proposed to solve SC is very large, see for instance~\cite{foucart2013introduction} for a comprehensive overview. Overall, there exist at least three kind of heuristics to provide candidate solutions to SC: 
\begin{itemize}
  \item Greedy methods, such as OMP~\cite{Mallat1993Matching,Pati1993Orthogonal} which is described in more details in Section~\ref{sec:HOMP}. These methods select indices where \(x\) is nonzero greedily until the target sparsity level or a tolerance on the reconstruction error is reached. They benefit from optimality guarantees when the dictionary columns, called atoms, are ``far'' from each others. In that case the dictionary is said to be incoherent~\cite{Tropp2004Greed}.
  \item Nonconvex first order methods, such as IHT~\cite{blumensath2008iterative}, that are based on proximal gradient descent~\cite{Parikh2014Proximal} knowing that the projection on the set of \(k\)-sparse vectors is obtained by simply clipping the \(n-k\) smallest absolute values in \(x\). Convergence guarantees for first order constrained optimization methods is a rapidly evolving topic out of the scope of this communication, but a discussion on optimality guarantees of IHT is available in Section~\ref{sec:iht}.
  \item Convex relaxation methods, such as Least Absolute Shrinkage and Selection Operation (LASSO). The non-convex \(\ell_0\) constraint in Equation~\ref{eq:SC} is replaced by a surrogate convex constraint, typically using the \(\ell_1\) norm. This makes the problem convex and easier to solve, at the cost of potentially changing the support of the solution~\cite{Tropp2006Just}. Solving the LASSO can be tackled with the Fast Iterative Soft Thresholding Algorithm (FISTA)~\cite{Nesterov1983method,Beck2009Fast} which is essentially an accelerated proximal gradient method with convergence guarantees.
\end{itemize}

\subsection{Models closely related to DLRA}

While this work is interested in constraining the factor matrix \(A\) in Equation~\eqref{eq:LRA} so that its columns are sparse in a known dictionary (in other words, sparse coding the columns of \(A\)), a few previous works have been concerned with encoding each such column with only one atom. Most related to this work is the so-called Dictionary-based CPD~\cite{Cohen2018Dictionary}, which I shall rename as one-sparse dictionary-based LRA. Within this framework, a model such as NMF becomes
\begin{equation}\label{eq:1sNMF}
  \underset{\mathcal{K}\in\mathcal{P}_r([1,d]),\; B\geq 0}{\argmin}~\|Y - D_{\mathcal{K}}B^T \|_F^2
\end{equation}
where \(\mathcal{P}_r([1,d])\) is the set of all parts of \([1,d]\) with \(r\) elements, therefore \(\mathcal{K}\) is a set of \(r\) indices from \(1\) to \(d\). This problem is a particular case of the proposed DLRA framework because \(D_{\mathcal{K}}\) can be written as \(DX\) where the columns of \(X\) are one-sparse. It was shown that one-sparse DLRA makes low-rank matrix factorization identifiable under mild coherence conditions on the dictionary, and makes the computation of CPD a well-posed problem. Intuitively, the dictionary constraint may be used to enforce a set of known templates to be used as patterns in the pattern mining procedure, and therefore one-sparse DLRA may be seen as a glorified pattern matching technique.

Other works in the sparse coding literature are related to DLRA, in particular the so-called multiple measurement vectors or collaborative sparse coding~\cite{cotter2005sparse,Elhamifar2012See,Iordache2014Collaborative}, which extends sparse coding when several inputs collected in a matrix \(Y\) are coded with the same dictionary using the same support. Effectively this means solving a problem of the form
\begin{equation}
  \underset{\|\sum_{i=1}^{r} Z_{i}^2\|_{0}\leq kr}{\argmin}~\| Y - DZ^T\|_F^2
\end{equation}
where the square is meant elementwise. DLRA can also be seen as a collaborative sparse coding by noticing that \(Z:=XB^T\) is at least \(kr\) row-sparse in Equation~\eqref{eq:DLRA}. However the low-rank hypothesis is lost, as well as the affectation of at most \(k\) atoms per column of matrix \(A\).

\section{Mixed Sparse Coding heuristics}

\subsection{Properties of Mixed Sparse Coding}\label{sec:propmsc}

In order to design efficient heuristics to solve MSC, let us first cover a few properties of MSC such as uniqueness of MSC solutions, relate MSC with other sparse coding problems, and check whether special simpler cases exists such as when support of the solution is known or when the dictionary is orthogonal. The following section covers this material.

\subsubsection{Equivalent formulations and relation to sparse coding}

Let us start our study by linking MSC to other sparse coding problems. For reference sake, in this work the standard matrix sparse coding problem is formulated as
\begin{equation}\label{eq:mSC}
  \underset{\forall i\leq r,\;\|X_i\|_0\leq k}{\argmin} \|Y - DX\|_F^2 ~.
\end{equation}
which is equivalent to solving \(r\) vector sparse coding problems as defined in Equation~\eqref{eq:SC}.

Because the Frobenius norm is invariant to a reordering of the entries, it can be seen easily that Problem~\eqref{eq:MSC} is equivalent after vectorization\footnote{Vectorization in this manuscript is row-first~\citep{Cohen2015About}} to a structured vector sparse coding Problem~\eqref{eq:SC} with block-sparsity constraints:
\begin{equation}\label{eq:vecpb}
  \underset{ \forall i\leq r,\; \|\vecn(X_i)\|_0\leq k}{\argmin} \| \vecn(Y) - (D\otimes B)\left[\vecn(X_1), \ldots, \vecn(X_r)\right] \|_2^2~.
\end{equation}
It appears that MSC is therefore a structured sparse coding problem since the dictionary is a Kronecker product of two matrices. Moreover, the sparsity constraint applies on blocks of coordinates in the vectorized input \( \vecn(X) \), as studied in~\citep{traonmilin2018stable}. To the best of the author's knowledge, block sparse coding with Kronecker structured dictionary have not been specifically studied. In particular, the conjunction of Kronecker structured dictionary~\citep{Tsiligkaridis2013Covariance} and structured sparsity leads to specific heuristics detailed in the rest of this work.
Nevertheless, in Section~\ref{sec:ncvxheur}, it is shown that MSC reduces to columnwise sparse coding under a small noise regime.

On a different note, consider the following problem
\begin{equation}\label{eq:MSC2}
  \underset{\|Y-DXB^T\|^2_F\leq \epsilon_1}{\min} \max_i\|X_i\|_0~.
\end{equation}
for some positive constant \( \epsilon_1 \).
Just like how Quadratically Constrained Basis Pursuit and LASSO are equivalent~\citep{foucart2013introduction}, one can show using similar arguments that Problems~\eqref{eq:MSC} and~\eqref{eq:MSC2} have the same solution given a mild uniqueness assumption.

\begin{proposition}
    Suppose that Problem~\eqref{eq:MSC2} has a unique solution \( X^\ast \) with \( \epsilon_1\geq 0 \). Then there exist a particular instance of Problem~\eqref{eq:MSC} such that \( X^\ast \) is the unique solution of~\eqref{eq:MSC}. Conversely, a unique solution to Problem~\eqref{eq:MSC} is the unique solution to a particular instance of Problem~\eqref{eq:MSC2}.
\end{proposition}

Formulation~\eqref{eq:MSC2} of MSC could also be expressed with matrix induced norms. Indeed, defining \( \ell_{0,0}(X):= \text{sup}_{\|z\|_0=1} \|Xz\|_0 = \max_i \|X_i\|_0 \) as an extension of matrix induced norms\footnote{Mind the possible confusion with the \( \mathcal{L}_{p,q} \) convention \( \ell_{0,\infty} \) commonly encountered.} \( \ell_{p,q}(X):= \text{sup}_{\|z\|_p = 1}\|Xz\|_q \)~\citep{Golub1989Matrix}, Problem~\eqref{eq:MSC2} is equivalent to
\begin{equation}\label{eq:MSC2bis}
  \underset{\|Y-DXB^T\|^2_F\leq \epsilon_1}{\min} \ell_{0,0}(X)~,
\end{equation}
a clear structured extension of quadratically constrained sparse coding~\citep{foucart2013introduction}. Some authors preferably work with Problem~\eqref{eq:MSC2} rather than~\eqref{eq:MSC} because in specific applications, choosing an error tolerance \( \epsilon_1 \) is more natural than choosing the sparsity level \( k \). While heuristics introduced further are geared towards solving Problem~\eqref{eq:MSC}, they can be adapted to solve Problem~\eqref{eq:MSC2}.

The quadratically constrained formulation also sheds light on the fact that, if there exist some solution \( X \) to MSC such that the residual \( \|Y - DXB^T \|_F^2 \) is zero, then MSC is equivalent to matrix sparse coding Problem~\eqref{eq:mSC}. Indeed, since it is assumed that \( B \) is full column rank, any such solution yields \( YB(B^TB)^{-1} = DX \). Consequently, noiseless formulations of MSC will not be considered any further.

\subsubsection{Generic uniqueness of solutions to MSC}

In sparse coding, sparsity is introduced as a regularizer to enforce uniqueness of the regression solution. It is therefore natural to wonder if this property also holds for MSC. It is not difficult to observe that indeed generically the solution to MSC will be unique, under the usual spark condition on the dictionary \( D \)~\citep{donoho2003optimally}.

\begin{proposition}
  Define \( \sparkn(D) \) as the smallest number of columns of \( D \) that are linearly dependent. Suppose that \( \sparkn(D)>2k \) and suppose \( B \) is a full column rank matrix. Then the set of \( Y \) such that Problem~\eqref{eq:MSC} has strictly more than one solution has zero Lebesgue measure.
\end{proposition}


This result basically states that in practice, most MSC instances will have unique solutions as long as the dictionary is not too coherent.

\subsubsection{Solving MSC exactly when the support is known}\label{sec:fixedsupp}
Solving the NP-hard MSC problem exactly is difficult because the naive, brute force algorithm implies testing all combinations of supports for all columns, which means computing \( {k\choose d}^r \) least squares and is significantly slower than brute force for Problem~\eqref{eq:mSC}. However, in the spirit of sparse coding which boils down to finding the optimal support for a sparse solution, MSC also reduces to a least squares problem when the locations of the zeros in a solution \( X \) are fixed and known. Below is detailed how to process this least squares problem to avoid forming the Kronecker matrix \( D\otimes B \) explicitly.

From the observation in Equation~\eqref{eq:vecpb} that the vectorized problem is really a structured sparse coding problem, one can check that for a given support \(S=\Supp(\vecn(X))\), solving MSC amounts to solving a linear system
\begin{equation}\label{eq:linsys}
  \underset{z\in\mathbb{R}^{kr}}{\argmin}~ \|y - {(D\otimes B)}_S z \|_2^2~.
\end{equation}
where \( y:=\vecn(Y) \).

It is not actually required to form the full Kronecker product \( D\otimes B \) and then select a subset \( S \) of its columns, nor is it necessary to vectorize \( Y \). 
More precisely, one needs to compute two quantities: \( \left((D\otimes B)_S\right)^T(D\otimes B)_S \) and \( \left((D\otimes B)_S\right)^T y\) to feed a linear system solver, and these products can be computed efficiently. Denote \( S_i \) the support of column \( X_i \). Formally, one may notice that \((D\otimes B)_S \) is a block matrix \([D_{S_1}\otimes b_1, \ldots, D_{S_r}\otimes b_r]\).
Therefore, using the identity \( (D\otimes b)^T(D\otimes b) = b^Tb D^TD \), one may compute the cross product by first precomputing \(U =  D^TD\)\footnote{If this is not possible because \(D\) is too large, one may instead compute the blocks \( U_{S_iS_j} \)without precomputing \( D^TD \). } and \(V = B^TB\), then computing each block \((D_{S_i}\otimes b_i)^T(D_{S_j}\otimes b_j) \) as \(   v_{ij} U_{S_iS_j} \).

For the data and mixing matrix product, notice that \( (D_{S_i}\otimes b_i)^Ty  = D_{S_i}Yb_i^T\). Therefore, the model-data product can be obtained by first precomputing \(N = YB^T\), then computing each block \( \left((D\otimes B)_{S_i}\right)^T y\) as \( D_{S_i} N_i\).

\underline{Remark on regularization}: It may happen that for a fixed support, the linear system~\eqref{eq:linsys} is ill-posed. This may be caused by a highly coherent dictionary \( D \) or the choice of a large rank \( r \). To avoid this issue, when the system is ill-conditioned in later experiments, a small ridge penalization may be added.
\subsubsection{MSC with orthogonal dictionary is easy for rank one LRA}\label{sec:ortho}

An important question is whether the problem generally becomes easier if the dictionary is left orthogonal. This is the case for Problem~\eqref{eq:mSC}, where orthogonal \( D \) allows to solve the problem using only Hard Thresholding (HT) on \( D^TY \) columnwise. Below, it is shown that a similar HT procedure can be used when \( r=1 \), but not in general when \( r>1 \).

Supposing \( D \) is left orthogonal, the MSC problem becomes
\begin{equation}\label{eq:orthoDmsc}
  \underset{\|X_i\|_0\leq k\;\forall i\in[1,r]}{\argmin}\|D^TY - XB^T \|_F^2~.
\end{equation}
When \( r=1 \), matrix \( B^T \) is simply a row vector, which is a right orthogonal matrix after \( \ell_2 \) normalization such that solving Problem~\eqref{eq:orthoDmsc} amounts to minimizing \( \|\frac{1}{\|b\|^2_2}D^TYb - x \|_2^2 \). Then the solution is obtained using the thresholding operator \( \text{HT}_k(\frac{1}{\|b\|^2_2}D^TYb) \) which selects the \( k \) largest entries of its input.

Sadly when \( r>1 \), matrix \( B^T \) is not right orthogonal in general and neither is \( D\otimes B \). Therefore this thresholding strategy does not yield a MSC solution in general despite \( D \) being left orthogonal.
\subsection{Non-convex Heuristics to solve Mixed Sparse Coding}\label{sec:ncvxheur}

We now focus on heuristics to find candidate solutions to MSC. Similarly to sparse coding, MSC is an NP-hard problem for which obtaining the global solution typically requires costly algorithms~\citep{bourguignon2015exact,nadisic2020exact}. Therefore, in the following section, several heuristics are proposed that aim at finding good sparse approximations in reasonable time.

\begin{itemize}
  \item A classical sparse coding algorithm (here OMP~\citep{Pati1993Orthogonal}) applied columnwise on \( YB(B^TB)^{-1} \). Indeed in a small noise regime, columnwise sparse coding on \( YB(B^TB)^{-1} \) is proven to be equivalent to solving MSC.
  \item A Block-coordinate descent algorithm that features OMP as a subroutine.
  \item A proximal gradient algorithm similar to Iterative Hard Thresholding.
  \item Two convex relaxations analogue to LASSO~\citep{tibshirani1996regression}, which are solved by accelerated proximal gradient in the spirit of FISTA~\citep{Beck2009Fast}. Maximum regularization levels are computed, and a few properties concerning the sparsity and the uniqueness of the solutions are provided.
\end{itemize}

The performance of all these heuristics as MSC solvers is studied in Section~\ref{sec:xpMSC}.


\subsubsection{A provable reduction to columnwise sparse coding for small noise regimes}\label{sec:TrickOMP}

At first glance, one may think that projecting \( Y \) onto the row-space of \( B^T \) maps solutions of Problem~\eqref{eq:MSC} to solutions of Problem~\eqref{eq:mSC}. Indeed, writing \( Y = Y_B + Y_{-B} \) with \( Y_{B}:=YB(B^TB)^{-1}B^T \) the orthogonal projection of \( Y \) on the row-space of \( B^T \), it holds that Problem~\eqref{eq:MSC} has the same minimizers than
\begin{equation}\label{eq:redMSC}
  \underset{\|X_i\|_0\leq k ~\forall i\in[1,r]}{\min} \|Y_B-DXB^T\|^2_F~.
\end{equation}
This problem is however not in general equivalent to matrix sparse coding because \( B^T \) distorts the error distribution.

\vspace{1em}
Even though they are not equivalent, one could hope to solve MSC, in particular settings, by using a candidate solution to the matrix sparse coding Problem~\eqref{eq:mSC} with \( YB(B^TB)^{-1} \) as input. The particular case of \(k=1\) is striking: matrix sparse coding is solved in closed form whereas MSC has no general solution as far as I know. This kind of heuristic replacement of Problem~\eqref{eq:MSC} by Problem~\eqref{eq:mSC} has been used heuristically in~\citep{Cohen2018Dictionary} when \( k=1 \), and can be interpreted as ``First find \( Z \) that minimizes \( \|Y - ZB^T \|_F^2 \), then find a \( k \)-sparse matrix \( X \) that minimizes \( \|Z - DX \|_F^2 \)''.

We show below that in a small perturbation regime, for a dictionary \( D \) and a mixing matrix \( B \) satisfying classical assumptions in compressive sensing, the matrix sparse coding solution has the same support than the MSC solution.

\begin{lemma}~\label{prop:robust}
  Let \( X,X' \) columnwise \( k \)-sparse matrices and \( \epsilon>0,\; \delta>0 \) such that \( \|Y-DXB\|_F^2 \leq \epsilon \) and \( \|YB(B^TB)^{-1} - DX' \|_F^2 \leq \delta \). Further suppose that \( \sparkn(D)>2k \) and that \( B \) has full column-rank. Then
  \begin{equation}
    \|X-X'\|_F \leq \frac{1}{\sigma^{(2k)}_{\min}(D)}\sqrt{\delta+\frac{\epsilon}{\sigma^2_{\min}(B)}}
  \end{equation}
  where \( \sigma_{\min}(B) \) is the smallest nonzero singular value of \( B \) and \( \sigma_{\min}^{(2k)} \) is the smallest nonzero singular value of all submatrices of \( D \) constructed with \( 2k \) columns.
\end{lemma}

Remarkably, if \( k=1 \), then Problem~\eqref{eq:mSC} has a closed-form solution, and one may replace \( \delta \) by the best residual to remove this unknown quantity in Lemma~\ref{prop:robust}. On a different note, variants of this result can be derived under similar assumptions, for instance if \( D \) satisfies a \( 2k \) restricted isometry property.

We can now derive a support recovery equivalence between MSC and SC.
\begin{proposition}
  Under the hypotheses of Lemma~\ref{prop:robust}, supposing that \( X \) and \( X' \) are exactly columnwise \( k \)-sparse, if
  \begin{equation}\label{eq:condxij}
    \min_{j\leq r}\sqrt{\min_{i\in \Supp(X_j)} {X_{ij}}^2 + \min_{ i'\in \Supp(X_j')} {X'_{i'j}}^2} >  \frac{1}{\sigma^{(2k)}_{\min}(D)}\sqrt{\delta+\frac{\epsilon}{\sigma^2_{\min}(B)}},
  \end{equation}
  then \( \Supp(X)=\Supp(X') \).
\end{proposition}
In practice, this bound can be used to grossly check whether support estimation in MSC may reduce to support estimation in matrix sparse coding or not. To do so, one may first tentatively solve Problem~\eqref{eq:mSC} with \( YB(B^TB)^{-1} \) as input and obtain a candidate value \( X' \) as well as some residual \( \delta \). Furthermore, while this is costly, the values of \( \sigma_{\min}(B) \) and \( \sigma_{\min}^{(2k)}(D) \) may be computed. Then removing the term \( X_{ij} \) from Equation~\eqref{eq:condxij} yields a bound of the noise level \( \epsilon \):
\begin{equation}
  \sigma_{\min}(B)\left(-\delta + {(\sigma_{\min}^{(2k)}(D))}^2\min_{j, i\in \Supp(X')} {X'_{ij}}^2 \right) > \epsilon
\end{equation}
under which the reduction is well-grounded.

These observations lead to the design of a first heuristic, which applies OMP to each column of \( YB(B^TB)^{-1} \). The obtained support is then used to compute a solution \( X \) to MSC as described in Section~\ref{sec:fixedsupp}. This heuristic will be denoted TrickOMP in the following.

\subsubsection{A first order strategy: Iterative Hard Thresholding}\label{sec:iht}

Maybe the most simple way to solve MSC is by proximal gradient descent. In sparse coding, this type of algorithm is often referred to as Iterative Hard Thresolding (IHT), and this is how I will denote this algorithm as well for MSC.

Computing the gradient of the differentiable convex term \(f(X) =  \|Y - DXB \|_F^2 \) with respect to \( X \) is easy and yields
\begin{equation}
  \frac{\partial f}{2\partial X} =  - D^TYB + D^TDXBB^T~.
\end{equation}
Note that depending on the structure \( \Omega_B \), products \( D^TYB \) and \( B^TB \) may computed efficiently.

Then it is required to compute a projection on the set of columnwise \(k\)-sparse matrices. This can easily be done columnwise by hard thresholding. However, for the algorithm to be well-defined and deterministic, we need to suppose that projecting on the set of columnwise \( k \)-sparse  matrices is a closed-form single-valued operation. This holds if, when several solution exist, \textit{i.e.} when several entries are the \( k \)-th largest, one picks the right number of these entries in for instance the lexicographic order. The proposed IHT algorithm leverages the classic IHT algorithm and is summarized in the supplementary materials. However, contrarily to the usual IHT, the proposed implementation makes use of the inertial acceleration proposed in FISTA~\cite{Beck2009Fast}. Compared to using IHT for solving sparse coding, as hinted in Equation~\eqref{eq:vecpb}, here IHT is moreover applied to a structured sparse coding problem. This does not significantly modify the practical implementation of the algorithm.

IHT benefits from both convergence results as a (accelerated) proximal gradient algorithm with semi-algebraic regularization~\citep{blumensath2008iterative,attouch2013convergence} and support recovery guaranties for sparse coding~\citep{blumensath2009iterative,foucart2013introduction}. While the convergence results directly apply to MSC, extending support recovery guaranties to MSC is an interesting research avenue.

\subsubsection{Hierarchical OMP}\label{sec:HOMP}

Before describing the proposed hierarchical algorithm, it might be helpful to understand how greedy heuristics for sparse coding, such as the OMP algorithm, are derived. Greedy heuristics select the best atom to reconstruct the data, then remove its contribution and repeats this process with the residuals. The key to understanding these techniques is that selecting only one atom is a problem with a closed-form solution. Indeed, for any \( y\in\mathbb{R}^{n} \),
\begin{equation}
  \underset{\|x\|_0\leq 1}{\argmin}\|y - Dx\|_2^2 = \underset{z\in\mathds{R}}{\argmin}\underset{j\in[1,d]}{\argmin} \| y - zD_j\|_2^2
\end{equation}
and after some algebra, for a dictionary normalized columnwise,
\begin{equation}
  \|y - zD_j\|_2^2 = \|y \|_2^2 + z^2 -2 z {D_j^T}y = cst(j) -2 z{D_j^T}y
\end{equation}
which minimum with respect to \(j\) does not depend on the value \(z\) of the
nonzero coefficient \( z \) in the vector \(x\) (except for the sign of \(z\)).
Therefore the support of \(x\) is \(\text{argmax}_j |{D_j^T}y|\).

Reasoning in the same manner for MSC
does not yield a similar simple solution for finding the support. Indeed, even
setting \( k=1 \),
\begin{equation}
  \underset{\|X_i\|_0\leq 1}{\argmin}\|Y - DXB^T\|_F^2 = \underset{\forall i\leq r,\;z_i\in\mathds{R}}{\argmin}~~\underset{\forall i\leq r,\;j_i\in[1,d]}{\argmin} \| Y - \sum_{i=1}^{r} z_i D_{j_i} B_i^T\|_F^2
\end{equation}
and the interior minimization problem of the right-hand side is still difficult, referred to as dictionary-based low-rank matrix factorization in~\citep{Cohen2018Dictionary}. Indeed, the cost can be rewritten as \( \|Y - D(:,\mathcal{K}) Diag(z) B^T \|_F^2 \) with \( \mathcal{K} \) the list of selected atoms in \( D \). As discussed in Section~\ref{sec:TrickOMP}, the solutions to this problem in a noisy setting are in general not obtained by minimizing instead \( \|YB(B^TB)^{-1} - D(:,\mathcal{K})Diag(z) \|_F^2 \) with respect to \( \mathcal{K} \) which would be solved in closed form.

Therefore, even with \( k=1 \), a solution to MSC is not straightforward.
While a greedy selection heuristic may not be straightforward to design, one may notice that if the rank had been set to one, \textit{i.e.} if \( r=1 \), then for any \( k \) one actually ends up with the usual sparse coding problem.

Indeed, given a matrix \( V\in\mathbb{R}^{n\times m} \), the rank-one case means we try to solve the problem
\begin{equation}\label{eq:pb5}
  \underset{x\in\mathds{R}^{d}, \|x\|_0\leq k}{\argmin} \|V - Dxb^T \|_F^2
\end{equation}
which is equivalent to
\begin{equation}\label{eq:pb5bis}
  \underset{x\in\mathds{R}^{d}, \|x\|_0\leq k}{\argmin} \|\frac{1}{\|b\|_2^2}Vb - Dx \|_2^2~.
\end{equation}
This is nothing more than Problem~\eqref{eq:SC}. Consequently, we are now ready to design a hierarchical, \textit{i.e.} block-coordinate, greedy algorithm. The algorithm updates one column of \( X \) at a time, fixing all the others. Then, finding the optimal solution for that single column is exactly a sparse-coding problem.

This leads to an adaptation of OMP for MSC that I call Hierarchical OMP (HOMP), where OMP is used to solve each sparse coding subproblem, see
Algorithm~\ref{alg:HOMP}. The routine OMP\((x,D,k)\) applies Orthogonal Matching Pursuit to the input vector \( x \) with normalized dictionary \( D \) and sparsity level \( k \) and returns estimated codes and support. Note that after HOMP has stopped, it is useful to run a least square joint final update with fixed support as described in Section~\ref{sec:fixedsupp} because the final HOMP estimates may not be optimal for the output support.
HOMP does not easily inherit from OMP recovery conditions~\cite{Tropp2004Greed} because it employs OMP inside an alternating algorithm. On a practical side, any optimized implementation of OMP such as batch OMP~\cite{Rubinstein2008Efficient} can be used to implement HOMP as a subroutine.

\textbf{A note on restart:} A restart condition is required, \textit{i.e.} checking if the error increases after an inner iteration and rejecting the update in that case. Indeed OMP is simply a heuristic which, in general, is not guarantied to find the best solution to the sparse coding subproblem. When restart occurs, simply compute the best update with respect to the previously known support and move to the next column update. If restart occurs on all modes, then the algorithm stops with a warning. Due to this restart condition, the cost always decreases after each iteration, therefore it is guarantied that the HOMP algorithm either converges or stops with a warning.

\begin{algorithm}
  \caption{Hierarchical OMP}\label{alg:HOMP}
  \begin{algorithmic}
   \STATE{\textbf{Input:} data \( Y \), dictionary \( D \), sparsity level \( k \), initial value \( X \).}
   \STATE{\textbf{Output:} estimated codes \( X \), support \( S \)}
   \STATE{Precompute \( {D^T}D \) and \( {D^T}Y \) if memory allows.}
   \WHILE{ stopping criterion is not reached }
    \FOR{ \( p \) from \( 1 \) to \( r \)}
      \STATE{Set \( Vb= \frac{1}{\|B_p\|_2^2}(Y - DX_{-p}B^T_{-p})B_p \).}
      \STATE{Compute \( X_p,S_p = \text{OMP}(Vb,D,k) \)}
      \STATE{If error increased, reject this update, and perform a least squares update with the previous support. End if rejection happened for each column.}
    \ENDFOR{}
   \ENDWHILE{}
   \STATE{Set \( X \) as the least squares solution following Section~\ref{sec:fixedsupp} with support \( S \).}
  \end{algorithmic}
\end{algorithm}

\subsection{Convex heuristics to solve MSC}\label{sec:cvxheur}

The proposed greedy strategies TrickOMP and HOMP may not provide the best solutions to MSC or even converge to a critical point, and the practical performance of IHT is often not satisfactory, see Section~\ref{sec:xpMSC}.
Therefore, taking inspiration from existing works on sparse coding, one might as well tackle a convex problem which solutions are provably sparse. This means, first of all, finding convex relaxations to the \( \ell_{0,0} \) regularizer.

In what follows, we study two convex relaxations and propose a FISTA-like algorithm for each. In both cases,
the solution is provably sparse and there exist a regularization level such that the only solution is zero. Therefore, these regularizers force the presence of zeros in the columns of the solution.

\subsubsection{A columnwise convex relaxation: the Block LASSO heuristic}\label{sec:BlockLASSO}

A first convex relaxation of MSC is obtained by replacing each sparsity constraint \( \|X_i\|_0\leq k \) by an independent \( \ell_1 \) regularization. This idea has already been theoretically explored in the literature, in particular in~\citep{traonmilin2018stable} where several support recovery results are established.

Practically, fixing a collection \( \{\lambda_i\}_{i\leq r} \) of positive regularization parameters, the following convex relaxed problem, coined Block LASSO,
\begin{equation}\label{eq:pbcvx1}\tag{CVX1}
    \underset{X\in\mathds{R}^{d\times r}}{\argmin} \frac{1}{2}\|Y-DXB^T\|_F^2 + \sum_{i=1}^{r} \lambda_i \|X_i\|_1
\end{equation}
provides candidates solutions to MSC. This is a convex problem since each term is convex. Moreover the cost is coercive so that a solution always exists. However it is not strictly convex in general, thus several solutions may co-exist.

Adapting the proof in~\citep{foucart2013introduction}, one can easily show that under uniqueness assumptions, solutions to Problem~\eqref{eq:pbcvx1} are indeed sparse.
\begin{proposition}\label{prop:BLASSOsparse}
  Let \( X^\ast \) a solution to Problem~\eqref{eq:pbcvx1}, and suppose that \( X^\ast \) is unique. Then denoting \( S_i \) the support of each column \( X^\ast_i \), it holds that \( D_{S_i} \) is full column-rank, and that \( |S_i| \leq n \).
\end{proposition}
This shows that the Block LASSO solutions have at most \( n \) nonzeros in each column. Therefore, the columnwise \( \ell_1 \) regularization induces sparsity in all columns of \( X \) and solving Problem~\eqref{eq:pbcvx1} is relevant to perform MSC. But this does not show that the support recovered by solving Block LASSO is always the support of the solution of MSC, see~\citep{traonmilin2018stable} for such recovery results given assumptions on the regularizations parameters \( \lambda_i \).


Conversely, it is of interest to know above which values of \( \lambda_i \) the solution \( X \) is null. Proposition~\ref{prop:maxlambdai} states that such
a columnwise maximum regularization can be defined. This can be used to set the \( \lambda_i \) regularization parameters individually given a percentage of \( \lambda_{i, \max} \) and provide a better intuition for choosing each \( \lambda_i \).
\begin{proposition}\label{prop:maxlambdai}
  A solution \( X^\ast \) of Problem~\eqref{eq:pbcvx1} satisfies \( X^\ast=0 \) if and only if for all \( i\leq r \), \( \lambda_i\geq \lambda_{i,\max} \) where \(\lambda_{i,\max} = \|{D^T}YB_i \|_{\infty} \) (element-wise absolute value maximum). Moreover, if \( X_i^\ast=0 \) is a column of a solution, then \( \lambda_i \geq \lambda_{i,\max}\).
\end{proposition}

\underline{Relation with other models}:
Vectorization easily transform the matrix Problem~\eqref{eq:pbcvx1} into a vector problem reminiscent of the LASSO. In fact, if the regularization parameters \( \lambda_i=\lambda \) are set equal, then Problem~\eqref{eq:pbcvx1} is nothing else than the LASSO. A related but different model is SLOPE~\citep{bogdan2015Slope}. Indeed, in SLOPE, it is possible to have a particular \( \lambda_{ji} \) for each \textbf{sorted} entry \( X_{\sigma_i(j)i} \), but in Problem~\eqref{eq:pbcvx1} the regularization parameters are fixed by blocks and the relative order of elements among the blocks changes in the admissible space. Thus the relaxed Problem~\eqref{eq:pbcvx1} cannot be expressed as a particular SLOPE problem.

\underline{Solving block LASSO with FISTA}:
A workhorse algorithm for solving the LASSO is FISTA~\citep{Beck2009Fast}.
Moreover, the regularization term is separable in \(X_i\), and the proximal operator for each separable term is well-known to be the soft-thresholding operator
\begin{equation}
  S_{\lambda_i}(x) = {\left[|x|-\lambda \right]}^+\text{sign}(x),\; S_{[\lambda_1,\ldots,\lambda_r]}(X) = [S_{\lambda_1}(X_1), \ldots, S_{\lambda_r}(X_r)]
\end{equation}
understood element-wise. Therefore, the FISTA algorithm can be directly leveraged to solve Problem~\eqref{eq:pbcvx1}, see Algorithm~\ref{alg:fista_bl} coined Block-FISTA hereafter. Convergence of the cost iterates of this proposed extrapolated proximal gradient method is ensured as soon as the gradient step is smaller than the inverse of the Lipschitz constant of the quadratic term, which is given by \({\sigma(D)}^2{\sigma(B)}^2\) with \( \sigma(M) \) the largest singular value of \( M \). The resulting FISTA algorithm is denoted as Block FISTA. Note that after Block FISTA returns a candidate solution, this solution's support is extracted, truncated to be of size \( k \) columnwise using hard thresholding, and an unbiased MSC solution is computed with that fixed support.

\begin{algorithm}
  \caption{FISTA for Block LASSO (Block-FISTA)}\label{alg:fista_bl}
  \begin{algorithmic}
   \STATE{\textbf{Input:} data \(Y\), dictionary \(D\), mixing matrix \(B\), regularization ratio \(\alpha\in[0,1]^r\), sparsity level \(k\), initial value \(X\).}
   \STATE{\textbf{Output:} estimated codes \( X \), support \( S \).}
   \STATE{Precompute \(D^TD, D^TYB\) and \( B^TB \) if memory allows.}
   \STATE{Compute \( \lambda_{i,\max} \) as in Proposition~\ref{prop:maxlambdai}, and set \( \lambda_i = \alpha_i\lambda_{i,\max}, \; \lambda = [\lambda_1,\ldots,\lambda_r
   ] \)}
   \STATE{Compute stepsize \( \eta = \frac{1}{\sigma(D)^{2}\sigma(B)^{2}} \)}
   \STATE{Initialize \(Z=X, \; \beta=1\).}
   \WHILE{stopping criterion is not reached}
    \STATE{\(X_{\text{old}}=X\)}
    \STATE{\(X = S_{\eta\lambda}\left( Z - \eta(D^TDZB^TB - D^TYB) \right)\)}
    \STATE{\(\beta_{\text{old}} = \beta\)}
    \STATE{\( \beta = \frac{1}{2}(1+\sqrt{1+4\beta^2})\)}
    \STATE{\( Z = X + \frac{\beta_{\text{old}}-1}{\beta} \left( X-X_{\text{old}} \right) \)}
   \ENDWHILE
   \STATE{Estimate the support \( S=S(X) \)}
   \STATE{Set \( X \) as the least squares solution following Section~\ref{sec:fixedsupp} with support \( S \).}
  \end{algorithmic}
\end{algorithm}

\subsubsection{Mixed \( \ell_1 \) norm for tightest convex relaxation}

The columnwise convex relaxation introduced in Section~\ref{sec:BlockLASSO} has the disadvantage of introducing a potentially large number of regularization parameters that must be controlled individually to obtain a target sparsity level columnwise. Moreover, this relaxation is not the tightest convex relaxation of the \( \ell_{0,0} \) regularizer on \( [-1,1] \).

It turns out that using the tightest convex relaxation of the \( \ell_{0,0} \) regularizer does solve the proliferation of regularization parameters problem, and in fact this yields a uniform regularization on the columns of \( X \). This comes however at the cost of loosing some sparsity guaranties as detailed below.

\begin{proposition}\label{prop:tight}
  The tightest convex relaxation of \(\ell_{0,0}\) on \( [-1,1]^{d\times r} \) is \( \ell_{1,1}: X \mapsto \max_i \|X_i \|_1 =: \|X\|_{1,1} \).
\end{proposition}

According to Proposition~\ref{prop:tight}, Problem~\eqref{eq:MSC} may be relaxed into the following convex optimization problem coined Mixed LASSO hereafter:
\begin{equation}\label{eq:conv_relax_unif}
  \underset{X\in\mathds{R}^{d\times r}}{\argmin} \frac{1}{2}\|Y-DXB^T\|_F^2 + \lambda \|X\|_{1,1}  ~.
\end{equation}

To again leverage FISTA to solve Problem~\eqref{eq:conv_relax_unif} and produce a support for a solution of MSC requires to compute the proximal operator of the regularization term. For the \( \ell_{1,1} \) norm, the proximal operator has been shown to be computable exactly with little cost using a bisection search~\citep{quattoni2009efficient,bejar2021fastest,cohen2020computing}. In this work I used the low-level implementation of~\citep{bejar2021fastest}\footnote{\url{https://github.com/bbejar/prox-l1oo}}. The resulting Mixed-FISTA algorithm is very similar to Algorithm~\ref{alg:fista_bl} but using the \( \ell_{1,1} \) proximal operator instead of soft-thresholding, and its pseudo-code is therefore differed to the supplementary materials.

\underline{Properties of the Mixed LASSO}:
Differently from the Block LASSO problem, the Mixed LASSO may not have sparse solutions if the regularization is not set high enough.
\begin{proposition}\label{prop:sparsitymLasso}
  Suppose there exist a unique solution \( X^{\ast} \) to the Mixed LASSO problem. Let \( \mathcal{I} \) the set of indices such that for all \( i \) in \( \mathcal{I}\), \(\|X^{\ast}_i\|_1 = \|X^{\ast}\|_{1,1} \). Denote \( S \) the support of \( X^{\ast} \). Then there exist at least one \( i \) in \( \mathcal{I} \) such that \( D_{S_i} \) is full column rank, and \( \|X^{\ast}_i \|_0 \leq n \). Moreover, if \( D \) is overcomplete, \( \mathcal{I}=\{1,\ldots,r\} \).
\end{proposition}
This result is actually quite unsatisfactory. The uniqueness condition, which is generally satisfied for the LASSO, seems a much stricter restriction in the Mixed LASSO problem. In particular all columns of the solution must have equal \( \ell_1 \) norm in the overcomplete case. Moreover sparsity is only ensured for one column. However, from the simulations in the Experiment section, it seems that in general the Mixed LASSO problem does yield sparser solutions than the above theory predicts. 

\underline{Maximum regularization}:
Intuitively, by setting the regularization parameter \( \lambda \) high enough, one expects the solution of the Mixed LASSO to be null. Below I show that this is indeed true.
\begin{proposition}\label{prop:maxlambda}
  \( X^{\ast}=0 \) is the unique solution to the Mixed LASSO problem if and only if \( \lambda \geq \lambda_{\text{max}} \) where \(\lambda_{\text{max}} =  \sum_{i=1}^{r} \|D^TYB_i\|_{\infty} \).
\end{proposition}

Similarly to Block-FISTA, this result can be used to chose the regularization parameter as a percentage of \( \lambda_{\text{max}} \).

\subsection{Nonnegative MSC}\label{sec:nnmsc}

In this section I quickly describe how to adapt the Block LASSO strategy in the presence of nonnegativity constraints on \( X \). This is a very useful special case of MSC since many LRA models use nonnegativity to enhance identifiability such as Nonnegative Matrix Factorization or Nonnegative Tucker Decomposition.

The nonnegative MSC problem can be written as follows:
\begin{equation}\label{eq:NNMSC}
  \underset{X\geq 0,\;\|X_i\|_0\leq k}{\argmin} \|Y - DXB^T\|_F^2 ~.
\end{equation}
Block LASSO minimizes a regularized cost which becomes quadratic and smooth due to the nonnegativity constraints:
\begin{equation}\label{eq:}
  \underset{X\geq 0}{\argmin} \|Y - DXB^T\|_F^2  + \sum_{i=1}^n \lambda_i 1^TX_i~.
\end{equation}
The Nonnegative least squares Problem~\eqref{eq:NNMSC} is easily solved by a modified nonnegative Block-FISTA, where the proximal operator is a projection on the nonnegative orthant.

While in the general case a least-squares update with fixed support is performed at the end of Block-FISTA to remove the bias induced by the convex penalty, in the nonnegative case a nonnegative least squares solver is used on the estimated support with a small ridge regularization. One drawback of this approach is that the final estimate for \( X \) might have strictly smaller sparsity level than the target \( k \) in a few columns.  
In the supplementary material, a quick comparison between Block-FISTA and its nonnegative variant shows the positive impact of accounting for nonnegativity for support recovery.

\subsection{Comparison of the proposed heuristics}\label{sec:xpMSC}

Numerical experiments discussed hereafter provide a first analysis of the proposed heuristics to solve MSC. A few natural questions arise upon studying these methods:
\begin{itemize}
  \item Are some of the proposed heuristics performing well or poorly in terms of support recovery in a variety of settings?
  \item Are some heuristics much faster than others in practice?
\end{itemize}
We shall provide tentative answers after conducting synthetic experiments. However, because of the variety of proposed methods and the large number of experimental parameters (dimensions, noise level, conditioning of \( B \) and coherence of \( D \), distribution of the true \( X \), regularization levels), it is virtually impossible to test out all possible combinations and the conclusions of this section can hardly be extrapolated outside our study cases. All the codes used in the experiments below are freely available online\footnote{\url{https://github.com/cohenjer/mscode} and \url{https://github.com/cohenjer/dlra}}. In particular, all the proposed algorithms are implemented in Python, and all experiments and figures can be reproduced using the distributed code.

In the supplementary material, I cover additional questions of importance: the sensitivity of convex relaxation methods to the choice of the regularization parameter, the sensitivity of all methods to the conditioning of \( B \) and the sensitivity to random and zero initializations.

\subsubsection{Synthetic experiments}

In general and unless specified otherwise we set \(n=50,m=50,d=100,k=5,r=6\). The variance of additive Gaussian white noise is tuned so that the empirical SNR is exactly \( 20\)dB. To generate \( D \), its entries are drawn independently from the Uniform distribution on \([0,1]\) and its columns are then normalized. The uniform distribution is meant to make atoms more correlated and therefore increase the dictionary coherence. The entries of \( B \) are drawn similarly. However, the singular value decomposition of \( B \) is then computed, its original singular values discarded and replaced with linearly spaced values from 1 to \(\frac{1}{\text{cond}(B)}\) where \(\text{cond}(B)=200\). The values of the true \( X \) are generated by first selecting a support randomly (uniformly), then sampling nonzero entries from standard Gaussian i.i.d. distributions.
The initial \( X \) is set to zero.

In most test settings, the metric used to assess performance is based on support recovery.
To measure support recovery, the number of correctly found nonzero position is divided by the total number of nonzeros to be found, yielding a \(100\%\) recovery rate if the support is perfectly estimated and \(0\%\) if no element in the support of \( X \) is correctly estimated.

The input regularization in Mixed-FISTA and Block-FISTA is always scaled from 0 to 1 by computing the maximum regularization, see Propositions~\ref{prop:maxlambdai} and~\ref{prop:maxlambda}. To choose the regularization ratio \( \alpha \), before running each experiment, Mixed-FISTA and Block-FISTA are ran on three instances of separately generated problems with the same parameters as the current test using a grid \([10^{-5},10^{-4},10^{-3},10^{-2},10^{-1}]\). Then the average best \( \alpha \) for these three tests is used as the regularization level for the whole test. This procedure is meant to mimic how a user would tune regularization for a given problem, generating a few instances by simulation and picking a vaguely adequate amount of regularization.

The stopping criterion for all methods is the same: when the relative decrease in cost \( \frac{|err^{it+1} - err^{it}|}{err^{it}} \) reaches \( 10^{-6} \), the algorithm stops. The absolute value allows for increasing the cost. Note that the cost includes the penalty terms for convex methods. Additionally, the maximum number of iterations is set to 1000.

\subsubsection{Test 1: support recovery vs. noise level}
\begin{figure}
  \includegraphics[width=\textwidth]{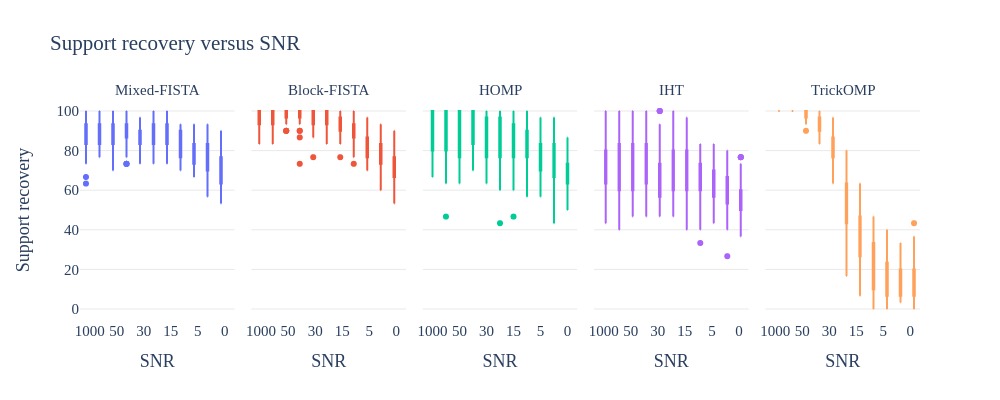}
  \caption{Support recovery (in \( \% \)) of the proposed heuristics to solve MSC at various Signal to Noise ratios. A total of \( 50 \) problems instances are used, with a single random but shared initialization for all methods.}
  \label{fig:Test1}
\end{figure}

For the first experiment, the noise level varies in power such that the SNR is exactly on a grid \( [1000, 100, 50, 40, 30, 20, 15, 10, 5, 2, 0] \). A total of \( 50 \) realizations of triplets \( (Y,D,B) \) are used in this experiment, as well as in Test 2 and 3. Results are shown in Figure~\ref{fig:Test1}.

It can be observed that IHT performs poorly for all noise levels. As expected, TrickOMP performs well only at low noise levels. HOMP and the FISTA methods have degraded performance when the SNR decreases, but provide with satisfactory results overall. Block-FISTA seems to perform the best overall.

\subsubsection{Test 2: support recovery vs. dimensions (k,d)}
\begin{figure}
  \includegraphics[width=\textwidth]{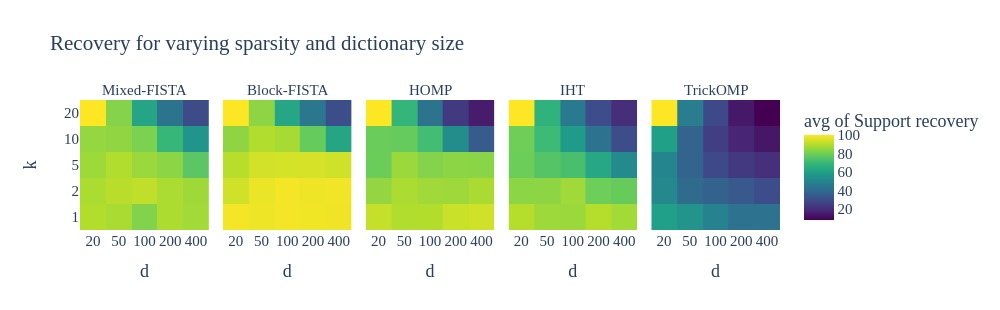}
  \caption{Support recovery (in \( \% \)) of the proposed heuristics to solve MSC for different sparsity levels \( k \) and different number of atoms \( d \) in the dictionary. A total of \( 50 \) problems instances are used, with a single random but shared initialization for all methods. When \( k=d=20 \), the support estimation is trivial thus all methods obtain \( 100\% \) support recovery.}
  \label{fig:Test2}
\end{figure}

This time the sparsity level \(k\) is on a grid \( [1, 2, 5, 10, 20] \) while the number of atoms \( d \) is also on a grid \( [20, 50, 100, 200, 400] \).
Figure~\ref{fig:Test2} shows the heat map results.

The TrickOMP has strikingly worse performance than the other methods. Moreover, as the number of atoms \( d \) increases or when the number of admissible supports \( d\choose k \), correct atom selection becomes more difficult for all methods. Again, Block-FISTA seems to perform better overall, in particular for large \( d \).

\subsubsection{Test 3: runtime vs. dimensions (n,m) and (k,d)}
In this last test, all algorithms are run until convergence for various sizes \( n=[10,50,1000] \) and \( m=[10,50,1000] \), or for various sparsity parameters \( k=[5,10,30] \) and \( [50,100,1000] \). Table~\ref{tab:Test3} provides runtime and number of iterations averaged for \( N=10 \) runs.


\begin{table}
  \footnotesize
  \caption{Average computation time in seconds and number of iterations with respect to \( (n,m) \) and \( (k,d) \). The format is (time, it). Computations are single threaded, and use a Intel® Core™ i7-8650U CPU @ 1.90GHz × 8 processor. TrickOMP exactly runs \( k \) steps of OMP equivalent to one HOMP inner iteration, therefore TrickOMP iterations are not reported in the table. Maximal number of iteration was set to 1000.}
  \label{tab:Test3}
  \begin{center}
  \begin{tabular}{ c||c|c|c|c|c|c|c|c|c| }
    (n,m)&(10,10)&(10,50)&(10,1000)& (50,10)&(50,50)&(50,1000)&(1000,10)&(1000,50)&(1000,1000)\\
    \hline
    HOMP & 0.8, 148 & 0.9, 149 & 2, 88 & 2.6, 450 & 2, 330 & 10, 290 & 13, 280 & 13, 247 & 70, 376 \\
    M.-FISTA & 0.4, 1000 & 0.4, 950 & 0.5, 875 & 0.4, 842 & 0.4, 799 & 1.6, 734 & 0.6, 393 & 1, 365 & 4.4, 388 \\
    B.-FISTA & 0.3, 998 & 0.3, 898 & 0.3, 907 & 0.3, 843 & 0.3, 764 & 1, 724 & 0.5, 376 & 0.7, 369 & 4.2, 388 \\
    IHT & 0.1, 356 & 0.2, 354 & 0.1, 254 & 0.07, 121 & 0.09, 105 & 0.2, 116 & 0.1, 92 & 0.2, 63 & 0.9, 80 \\
    TrickOMP & 0.03, -- & 0.03, -- & 0.03, -- & 0.03, -- & 0.03, -- & 0.04, -- & 0.08, -- & 0.07, -- & 0.09, --
  \end{tabular}
  \vspace{1cm}

  \begin{tabular}{ c||c|c|c|c|c|c|c|c|c| }
    (d,k)&(50,5)&(50,10)&(50,30)& (100,5)&(100,10)&(100,30)&(1000,5)&(1000,10)&(1000,30)\\
    \hline
    HOMP & 0.59, 95 & 2.2, 185 & 64, 886 & 0.39, 50 & 4.3, 345 & 14, 176 & 1.3, 40 & 3.9, 53 & 53, 161 \\
    M.-FISTA & 0.1, 356 & 0.3, 467 & 0.5, 456 & 0.37, 691 & 0.56, 814 & 0.78, 609 & 8.9, 1000 & 9.6, 1000 & 9.5, 1000 \\
    B.-FISTA & 0.12, 492 & 0.23, 319 & 0.57, 373 & 0.23, 668 & 0.38, 679 & 0.63, 619 & 5.2, 1000 & 4.7, 369 & 5.4, 1000 \\
    IHT & 0.05, 103 & 0.15, 92 & 0.69, 1000 & 0.06, 111 & 0.15, 100 & 0.52, 324 & 2.5, 215 & 3.3, 353 & 4.6, 528 \\
    TrickOMP & 0.02, -- & 0.08, -- & 0.57, -- & 0.43, -- & 0.14, -- & 0.55, -- & 0.04, -- & 0.13, -- & 0.67, --
  \end{tabular}
  \end{center}
\end{table}

From Table~\ref{tab:Test3}, it can be inferred that HOMP is often much slower than the other methods. TrickOMP is always very fast since it relies on OMP which runs in exactly \( k \) iterations. Block-FISTA generally runs faster than Mixed-FISTA. Moreover, it is not much slower than faster methods such as IHT and TrickOMP, in particular for larger sparsity values.

\subsubsection{Discussion}

In all the experiments conducted above and in the supplementary materials, the Block-FISTA algorithm provides the best trade-off between support recovery and computation time. Moreover, it is very easy to extend to nonnegative low-rank approximation models which are very common in practice. Therefore, to design an algorithm for DLRA, we shall make use of Block-FISTA (topped with a least-squares update with fixed support) as a solver for the MSC sub-problem. Note however than Block-FISTA required to select many regularization parameters, but the proposed heuristic using a fixed percentage of \( \lambda_{i,\max} \) worked well in the above experiments.

\section{Dictionary-based Low Rank Approximations}\label{part:2}

\subsection{A generic AO algorithm for DLRA}\label{sec:AODLRA}



Now that a reasonably good heuristic for solving MSC has been found, let us introduce an AO method for DLRA based on Block-FISTA. The proposed algorithm is coined AO-DLRA and is summarized in Algorithm~\ref{alg:aodlra}. It boils down to solving for \( X \) with Block-FISTA (Algorithm~\ref{alg:fista_bl}) and automatically computed regularization parameters, and then solving for the other blocks using any classical alternating method specific to the LRA at hand. Because solving exactly the MSC problem is difficult, even using Block-FISTA with well tuned regularization parameters, it is not guarantied that the \( X \) update will decrease the global cost. In fact in practice the cost may go up, and storing the best update along the iterations is good practice.

\begin{algorithm}
  \caption{An AO algorithm for DLRA (AO-DLRA)}\label{alg:aodlra}
  \begin{algorithmic}
   \STATE{\textbf{Input:} Initial guesses \( X^{(0)}, B^{(0)} \), data \( Y \), dictionary \( D \), sparsity level \( k\leq n \), initial regularization \( \alpha\in[0,1]^{r} \), iteration number \( l_{\max} \), sparsity tolerance \( \tau \).}
   \STATE{\textbf{Output:} Best estimated factors \( X^{(l)} \) and \( B^{(l)} \)}
   \STATE{Precompute \( D^TD, \;D^TY \) and \( \sigma_{D} = \sigma(D^TD) \) if memory allows.}
   \FOR{\( l = 1\ldots l_{\max} \)}
     \STATE{\underline{\( B \) update: }}
     \STATE{Compute \( B^{(l)}\in\Omega_B \) that decreases the cost in Problem~\eqref{eq:DLRA} with respect to \( B \).}
     \STATE{\underline{\(X\) update:}}
     \STATE{Stepsize evaluation: \( \eta^{(l)}= \frac{1}{\sigma_D\sigma({B^{(l)}}^TB^{(l)})} \).}
     \STATE{Compute efficiently data-factor product\( (D^TY)B^{(l)} \) and inner products \( {B^{(l)}}^TB^{(l)} \).}
     \STATE{($\ast$) Update \(X^{(l)}\) using Block-FISTA (Algorithm~\ref{alg:fista_bl}) with regularization parameters \( \alpha \), stepsize \( \eta^{(l)} \) and initial guess \( X^{(l-1)} \) }
     \WHILE{ \( \exists i\leq r,\; \| X^{(l)}_i \|_0 \notin [k, k+\tau] \)}
      \FOR{\( i = 1\ldots r \)}
        \IF{\( \| X^{(l)}_i \|_0 \leq k \)}
          \STATE{Decrease regularization \(\alpha := \alpha/1.3\) }
        \ELSIF{ \( \|X^{(l)}_i\|_0 \geq k+\tau\)}
          \STATE{Increase regularization \( \alpha := \min(1.01\alpha,1) \)}
        \ENDIF
      \ENDFOR
      \STATE{Go to ($\ast$) with \( X^{(l-1)}:=X^{(l)} \)}
     \ENDWHILE
     \STATE{Unbiaised estimation: update \( X^{(l)} \) with fixed support \( S_{X^{(l)}} \) as in Section~\ref{sec:fixedsupp}.}
     \STATE{Store \( X^{(l)} \) and \( B^{(l)} \) if residuals \( \|Y - DX^{(l)}B^{(l)}\|_F^2 \) have improved with respect to previous best.}
   \ENDFOR
  \end{algorithmic}
\end{algorithm}

\subsubsection{Selecting regularization parameters}

It had already been noted in Section~\ref{sec:xpMSC} that choosing the multiple regularization parameters \( \lambda_i \) of Block-FISTA can be challenging. In the context of Alternating Optimization, this is even more true. Indeed, the (structured) matrix \( B \) is updated at each outer iteration, therefore there is a scaling ambiguity between \( X \) and \( B \) that makes any arbitrary choice of regularization level \( \lambda_i \) meaningless. Moreover the values \( \lambda_{i,\max} \) change at each outer iteration. Consequently, obtaining a sparsity regularization percentage \( \alpha_i \) in each column of \( X \) at each iteration is challenging without some ad-hoc tuning in each outer iteration. To that end, the regularization percentages \(\alpha = [\alpha_1,\ldots,\alpha_r]\) are tuned inside each inner loop until the columnwise number of nonzeros reaches a target range \( [k, k+\tau] \) where \( \tau \geq 0 \) is user-defined.
This range is deliberately shifted to the right so that the each column does not have size strictly less than \( k \) nonzeros. Indeed, in that situation, a few atoms would have to be chosen arbitrarily during the unbiased estimation. More precisely, when a column has too many zeros, \( \alpha_i \) is divided by 1.3, while it is multiplied by 1.01 if it has few nonzeros. Note that an interesting research avenue would be to use an adaption of homotopy methods~\citep{osborne2000new} for Block LASSO instead of FISTA, which would remove the need for this heuristic tuning.

\subsubsection{A provably convergent algorithm: inertial Proximal Alternating Linear Minimization (iPALM)}\label{sec:ipalm}
 The AO-DLRA algorithm proposed above is a heuristic with several arbitrary choices and no convergence guaranties. If designing an efficient AO algorithm with convergence guarantees proved difficult, designing a convergent algorithm is in fact straightforward using standard block-coordinate non-convex methods. I focus in the following on the iPALM algorithm, which alternates between a proximal gradient step on \( X \) similar the one discussed in Section~\ref{sec:iht} and a proximal gradient step on \( B \). iPALM is guarantied to converge to a stationary point of the DLRA cost, despite the irregularity of the semi-algebraic \( \ell_{0,0} \) map. A pseudo-code for iPALM to initialize Algorithm~\ref{alg:aodlra} is provided in the supplementary materials.

\subsubsection{Initialization strategies}\label{sec:init}
Because DLRA is a highly non-convex problem, one can only hope to reach some stationary point of the cost in Problem~\eqref{eq:DLRA}. Furthermore,  the sparsity constraint on the columns of \( X \) implies that \( X_i \) must belong to a finite union of subspaces, making the problem combinatorial by nature. Using a local heuristic such as AO-DLRA or iPALM, it is expected to encounter many local minima --- a fact also confirmed in practical experiments reported in Section~\ref{sec:xpDLRA} and in previous works~\citep{Cohen2018Dictionary}. Therefore, providing an initial guess for \( X \) and \( B \) close to a good local minimum is important.

There are at least two reasonable strategies to initialize the DCPD model. First, for any low-rank approximation model which is mildly identifiable (such as NMF, CPD), the suggested method is to first compute the low-rank approximation with standard algorithms to estimate \( A^{(0)} \) and \( B^{(0)} \), and then perform sparse coding on the columns of \( A^{(0)} \) to estimate \( X^{(0)} \). The identifiability properties of these models should ensure that \( A^{(0)} \) is well approximated by \( DX \) with sparse \( X \). Second, several random initialization can be carried out, only to keep the best result.
A third option for AO-DLRA would be to use a few iterations of the iPALM algorithm itself initialized randomly, since iPALM iterations are relatively cheap. However it is shown in the experiments below that this method does not yield good results.

\subsection{Experiments for DLRA}\label{sec:xpDLRA}

In the next section, two DLRA models are showcased on synthetic and real-life data. First, the Dictionary-based Matrix Factorization (DMF, see below) model is explored for the task of matrix completion in remote sensing. It is shown that DMF allows to complete entirely missing rows, something that low-rank completion cannot do. Second, nonnegative DCPD (nnDCPD) is used for denoising smooth images in the context of chemometrics, and better denoising performance are obtained with nnDCPD than with plain nonnegative CPD (nnCPD) or when post-processing the results of nnCPD. Nevertheless, the goal of these experiments is not to establish a new state-of-the-art in these particular, well-studied applications, but rather to demonstrate the versatile problems that may be cast as DLRA and the efficiency of DLRA when compared to other low-rank strategies. 
The performance of AO-DLRA and iPALM in terms of support recovery and relative reconstruction error for DMF and DCPD is then further assessed on synthetic data.

\subsubsection{Dictionary-based matrix factorization with application to matrix completion}\label{sec:hsi-completion}

Let us study the following Dictionary-based Matrix Factorization model:
\begin{equation}\label{eq:DMF}
  \underset{X\in\mathbb{R}^{d\times r},\;\forall i\leq r, \; \|X_i\|_0\leq k,\;\; B\in\mathbb{R}^{m\times r}}{\argmin} \| Y-DXB^T \|_F^2~.
\end{equation}
Low-rank factorizations have been extensively used in machine learning for matrix completion~\citep{candes2009exact}, since the low-rank hypothesis serves as regularization for this otherwise ill-posed problem. A use case for DMF is the completion of a low-rank matrix which has missing rows. 
Using a simple low-rank factorization approach would fail since a missing row removes all information about the column-space on that row. Formally, if a data matrix \( Y \approx AB \) has missing rows indexed by \( I \), then the rows of matrix \( A \) in \( I \) cannot be estimated directly from \( Y \).
Dictionary-based low-rank matrix factorization circumvents this problem by expressing each column of \( A \) as a sparse combination of atoms in a dictionary \( D \), such that \( Y\approx DXB^T \) with \( X \) columnwise sparse. While fitting matrices \( X \) and \( B \) can be done on the known entries, the reconstruction \( DXB^T \) will provide an estimation of the whole data matrix \( Y \), including the missing rows. Formally, first solve
\begin{equation}\label{eq:dlra-missing-train}
  \underset{X\in\mathbb{R}^{(n-|I|)\times r},\;\forall i\leq r,\;\|X_i\|_0\leq k,\;B\in\mathbb{R}^{m\times r}}{\argmin} \|Y_{:,\overline{I}} - D_{:,\overline{I}}XB^T \|_F^2
\end{equation}
using Algorithm~\ref{alg:aodlra} and then build an estimation for the missing values in \( Y \) as \(\widehat{Y}_{:,I} = D_{:,I}XB^T \). Initialization is carried out using random factors sampled element-wise from a normal distribution.

In remotely acquired hyperspectral images, missing rows in the data matrix are common as they correspond to corrupted pixels. Moreover, it is well-known that many hyperspectral images are approximately low-rank. Therefore in this toy experiment, a small portion of the Urban hyperspectral image is used to showcase the proposed inpainting strategy. Urban is often considered to be between rank 4, 5 or 6~\citep{zhu2017hyperspectral}. I will use \( r=4 \) hereafter. Urban is a collection of 307x307 images collected on 162 clean bands. For the sake of simplicity, only a \(20 \times 20\) patch is considered, with 50 randomly-chosen pixels removed from this patch in all bands, and therefore after pixel vectorization the data matrix \( Y\in\mathbb{R}^{400\times 162} \) has 50 missing rows. Since columns of factor \( A \) in this factorization should stand for patches of abundance maps, they are reasonably sparse in a 2D-Discrete Cosine Transform dictionary \( D \), here using \( d=400 \) atoms.
Note that similar strategies for HSI denoising have been studied in the literature albeit without columnwise sparsity imposed on \( X \), see~\citep{zhuang2018fast}.

We measure the performance of two strategies: DMF computed on known pixels solving Problem~\eqref{eq:dlra-missing-train}, and OMP for each band individually. Both approaches use the same dictionary \( D \). To evaluate performance, the test estimation error on missing pixels \( \| Y_I - \widehat{Y}_I \|_F/\|Y_I\|_F \) is computed alongside with the average Spectral Angular Mapper (SAM)
\(\frac{1}{|I|}\sum_{i\in I} \arccos(\frac{Y_i^T \widehat{Y}_i}{\|Y_i\|_2\|\widehat{Y}_i\|_2}) \) on all missing pixels.
The sparsity level \( k \) is defined on the grid \( [10, 30, 50, 70, 100, 120, 150^*, 200^*, 250^*] \), where \( k^* \) is not used for the OMP inpainting because of memory issues. Results are averaged on \(N=20\) different initializations. For AO-DLRA, the initial regularization \( \alpha \) is set to \( 5\times 10^{-3} \), and \( \tau=20 \).

\begin{figure}
  \centering
  \includegraphics[width=0.329\textwidth]{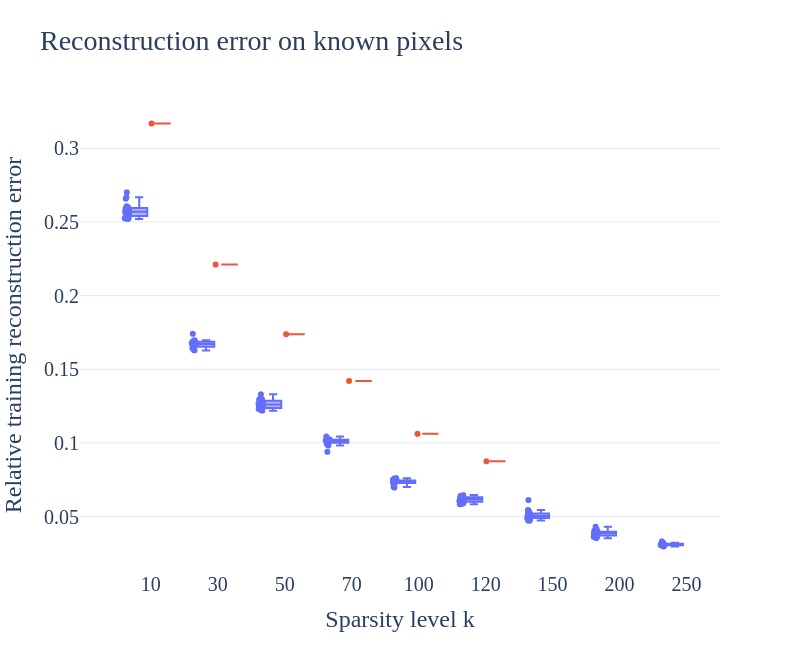}
  \includegraphics[width=0.329\textwidth]{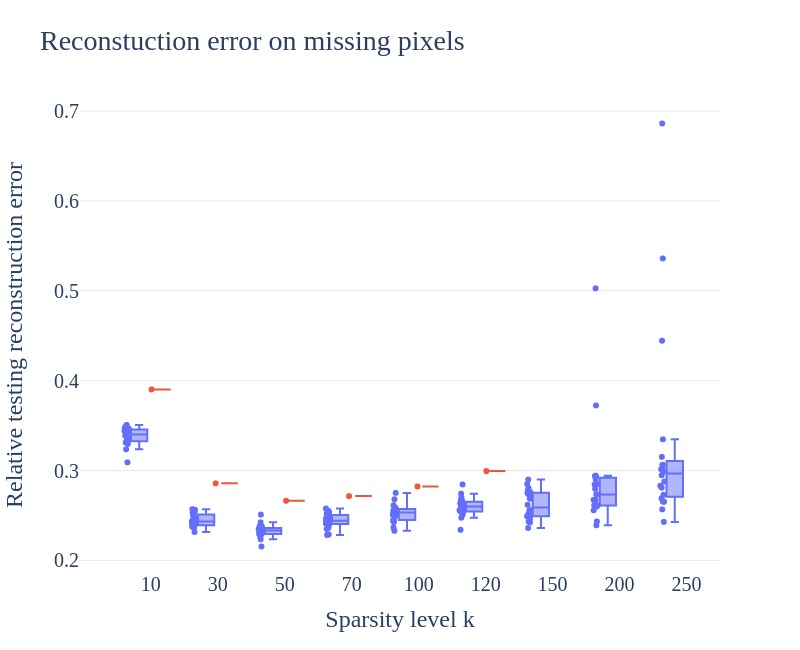}
  \includegraphics[width=0.329\textwidth]{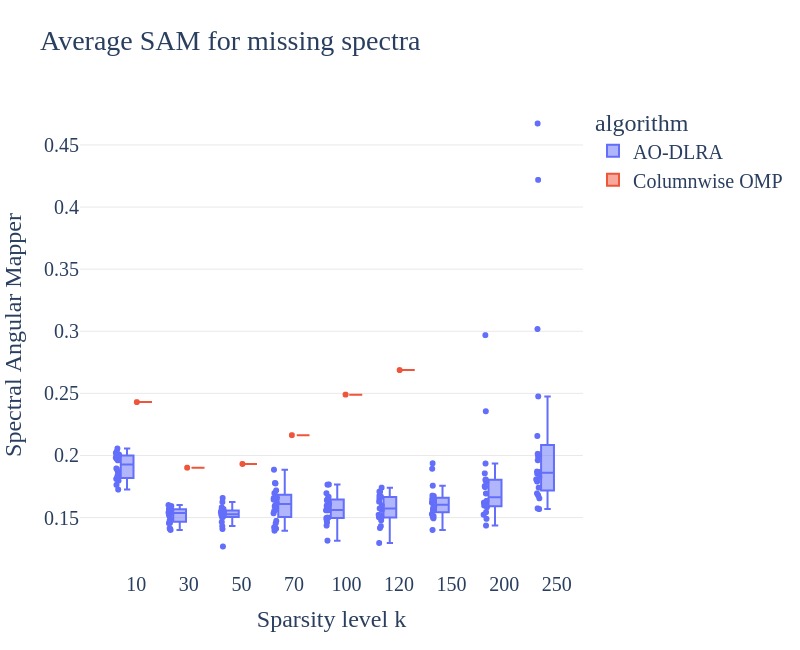}
  \caption{Performance of DMF (blue) versus Columnwise OMP (red, single valued box) for inpainting a patch of the Urban HSI with missing pixels at various sparsity levels. Twenty random initializations are used for DMF.}\label{fig:xpUrban}
\end{figure}

Figures~\ref{fig:xpUrban} shows the reconstruction error and SAM obtained after each initialization for various sparsity levels. First, for both metrics, there exist a clear advantage of the DMF approach when compared to sparse coding band per band. In particular, the band-wise OMP reconstruction does not yield good spectral reconstruction. Second, DMF apparently works similarly to sparse coding approaches for inpainting: if the sparsity level is too low the reconstruction is not precise, but if the sparsity level is too large the reconstruction is biased. Therefore, DMF effectively allows to perform inpainting with sparse coding on the factors of a low-rank matrix factorization.

\subsubsection{Dictionary-based Smooth Canonical Polyadic decomposition with application to data denoising}

The second study examines the DCPD discussed around Equation~\eqref{eq:DCPD} to perform denoising using smoothness. In this context, the data tensor \( Y \) is noisy, meaning formally that
\begin{equation}
  \tilde{Y} = Y + \epsilon, \;\text{and}\; Y = A(B\odot C)^T
\end{equation}
where \( \epsilon \) has a large power compared to \( A(B\odot C)^T \) (\textit{e.g.} Signal to Noise Ratio at -8.7dB in the following). Furthermore, let us suppose that \( A \) has smooth columns. The dictionary constraint can enforce smoothness on \( A \) by choosing \( D\) as a large collection of smooth atoms, in this case B-splines\footnote{The exact implementation of \( D \) is detailed in the code.}. Because the first mode is constrained such that \( A=DX \), each column of \( A \) is a sparse combination of smooth functions and will therefore itself be smooth. The sparsity constraint \( k \) prevents the use of too many splines and ensures that \( A \) is indeed smooth. Hereafter, the sparsity value is fixed to \( k=6 \).

There has been significant previous works on smooth CPD, perhaps most related to the proposed approach is the work of
\citep{Timmerman2002Three}. Their method also consists in choosing a dictionary \( D \) of B-splines. However, this dictionary has very few atoms (the actual number is determined by cross-validation), and picking the knots for the splines requires either time-consuming hand crafting, or some cross validation set. The advantage of their approach is that no sparsity constraint is imposed since there are already so few atoms in \( D \), and the problem becomes equivalent to CPD on the smoothed data.

The rationale however is that heavily crafting the splines is unnecessary. Using DCPD allows an automatic picking of good (if not best) B-splines. Furthermore, each component in the CPD may use different splines while the method of~\citep{Timmerman2002Three} uses the same splines for all the components. Hand-crafting is still required to build the dictionary, but one does not have to fear introducing an inadequate spline.

An advantage of B-splines is that they are nonnegative, therefore one can compute nonnegative DCPD by imposing nonnegativity on the sparse coefficients \( X \) as explained in Section~\ref{sec:nnmsc}. Imposing nonnegativity in the method of~\citep{Timmerman2002Three} is not as straightforward albeit doable~\citep{Cohen2015Fast}.

For this study the toy fluorescence spectroscopy dataset ``amino-acids'' available online\footnote{\url{http://www.models.life.ku.dk/Amino_Acid_fluo}} is used. Its rank is known to be \( r=3 \), and dimensions are \( 201\times 61 \times 5 \). Fluorescence spectroscopy tensors are nonnegative, low-rank and feature smooth factors. In fact factors are smooth on two modes related to excitation and emission spectra, therefore a double-constrained DCPD is also of interest. It boils down to solving
\begin{equation}\label{eq:D2CPD}
  \underset{\|X^{(A)}_i\|_0\leq k^{(A)},\; \|X_i^{(B)} \|_0\leq k^{(B)},\; C\in \mathbb{R}^{m_2\times r}}{\text{minimize}} \|\tilde{Y} - D^{(A)}X^{(A)}(D^{(B)}X^{(B)}\odot C)^T\|_F^2~.
\end{equation}
where \( D^{(A,B)} \) and \( k^{(A,B)} \) are the respective dictionaries of sizes \( 201 \times 180 \) and \( 61\times 81 \) and sparsity targets for modes one and two. The third mode factor contains relative concentrations~\citep{Bro1998Multi}. Additional Gaussian noise is added to the data so that the effective SNR used in the experiments is \( -8.7dB \). Because CPD is identifiable, in particular for the amino-acids dataset, nnCPD of the noisy data is used for initialization, computed using Hierarchical Alternating Least Squares~\citep{Phan2011Extended}. We compare sparse coding one or two modes of the output of HALS with AO-DCPD (\textit{i.e.} AO-DLRA used to compute DCPD) with smoothness on one mode, two modes, and on two modes with nonnegativity (AO-nnDCPD). We set \( \alpha=10^{-3} \) and \( \tau=5 \), and \( k^{(A)}=k^{(B)}=6 \).

\begin{figure}
  \centering
  \includegraphics[width=\textwidth]{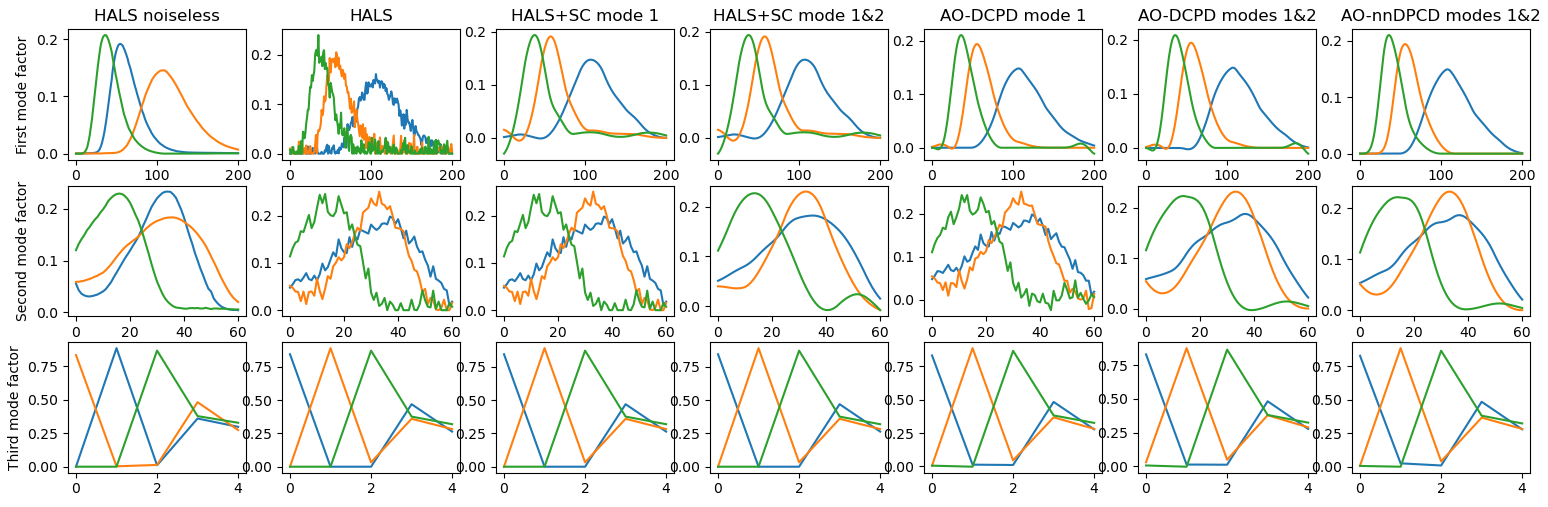}
  \caption{Factors estimated from nonnegative CPD and dictionary-based (nonnegative) CPD with smoothness imposed by B-splines on either one (emission, top row) or two (emission, excitation in middle row) modes. The third row shows the third mode factor that relates to relative concentration of the amino-acids in the mixture. The right-most plot shows AO-DLRA when imposing smoothness on two modes and nonnegativity on all modes.}
  \label{fig:chemo1}
\end{figure}

\begin{figure}
  \centering
  \includegraphics[width=\textwidth]{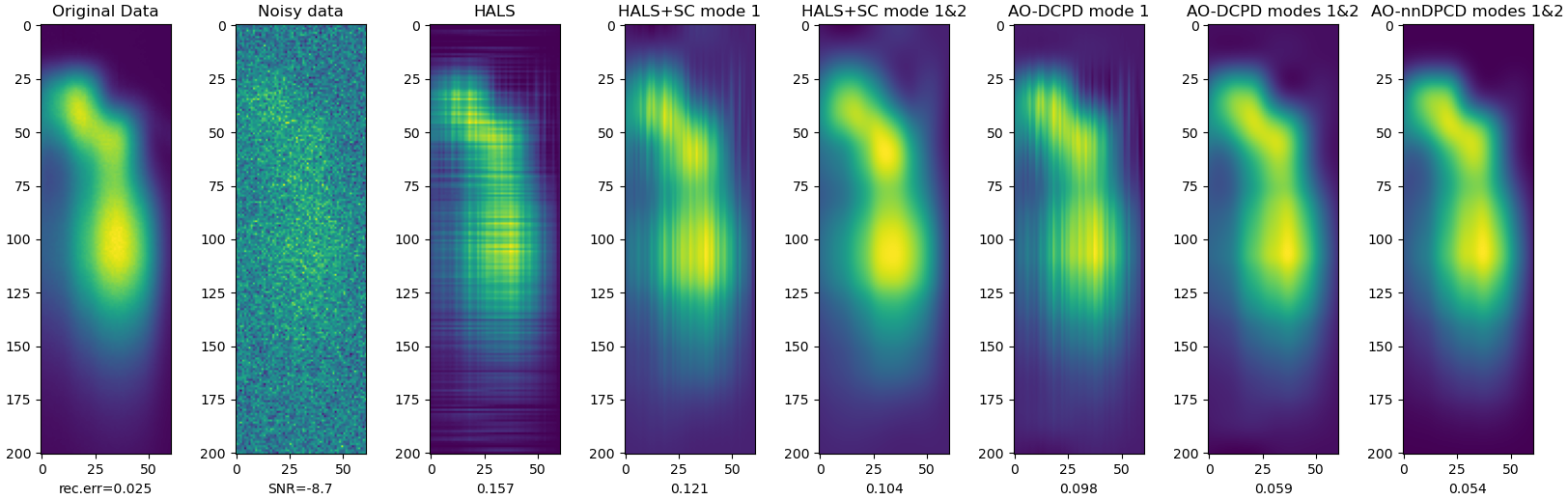}
  \caption{The fourth slice (index 3 if counting from 0) of the reconstructed tensors, see Figure~\ref{fig:chemo1} for details on each method. Row index relates to the emission wavelength while the column index relates to the excitation wavelength. The bottom values indicate the relative reconstruction error with respect to the clean tensor.}
  \label{fig:chemo2}
\end{figure}

Figure~\ref{fig:chemo1} shows the reconstructed factors and the relative error \( \|\widehat{Y} - Y \|_F/\|Y\|_F \) with respect to the true data, and Figure~\ref{fig:chemo2} shows one slice (here the fourth one) of the reconstructed tensor. It can be observed both graphically and numerically that DCPD techniques are overall superior to post-processing the output of HALS. In particular, the factors recovered using nnDCPD with dictionary constraints on two modes are very close to the true factors (obtained from the nnCPD of the clean data) using only \( k=6 \) splines at most.

\subsubsection{Performance of AO-DLRA for DMF and DCPD on synthetic data}

To assess the performance of AO-DLRA in computing DMF and DCPD, the support recovery and reconstruction error are monitored on synthetic data. For both model, a single initialization is performed for \( N=100 \) problem instances with the same hyperparameters. Matrices \( D,X,B,C \) involved in both problems and the noise tensors are generated as in Section~\ref{sec:xpMSC}. The rank is \( r=6 \) for a sparsity level of \( k=8 \) and the conditioning of \( B \) is set to \( 2\times10^{2} \). In the DMF experiment, the sizes are \( n=m=50 \), \( d=60 \) and the SNR is 100dB. For DPCD, we set \( n=20, m_1=21, m_2=22 \), \( d=30 \) and the SNR is 30dB. We set \( \alpha=10^{-2},\tau=20 \) for AO-DMF and \( \alpha=10^{-4}, \tau=20 \) for AO-DCPD.

A few strategies are compared: AO-DLRA initialized randomly, iPALM initialized randomly, and AO-DLRA initialized with iPALM. The same random initialization is used for all methods in each problem instance. For DCPD, we also consider AO-DLRA and iPALM initialized with a CPD solver (here Alternating Least Squares), and sparse coding the output of the ALS.  AO-DLRA always stops after at most 100 iterations, while iPALM runs for at most 1000 iterations or when relative error decrease is below \(10^{-8}\). The safeguard for stepsize in iPALM is set to \( \mu= 0.5\) for DMF and  \(\mu=1\) for DCPD.

\begin{figure}[H]
	\centering
	\includegraphics[width=0.49\textwidth]{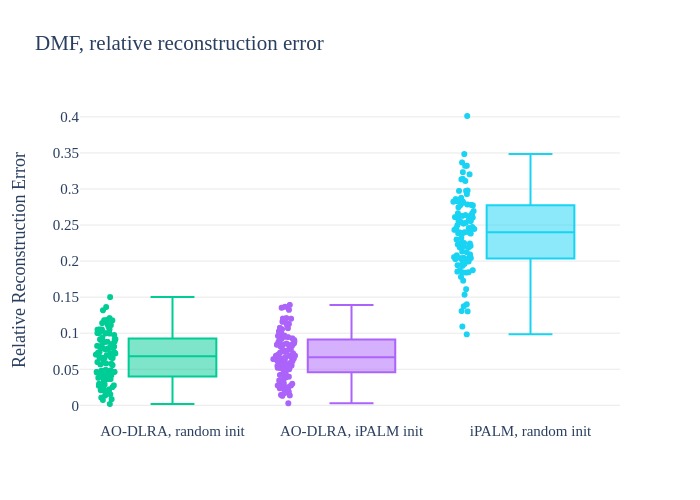}
	\includegraphics[width=0.49\textwidth]{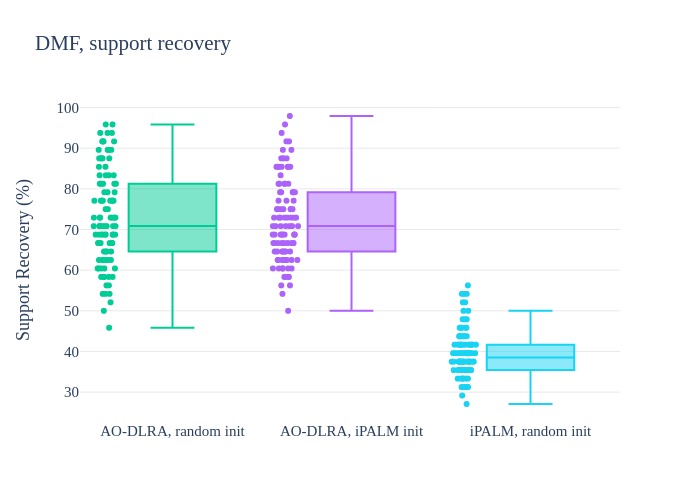}
	\includegraphics[width=0.49\textwidth]{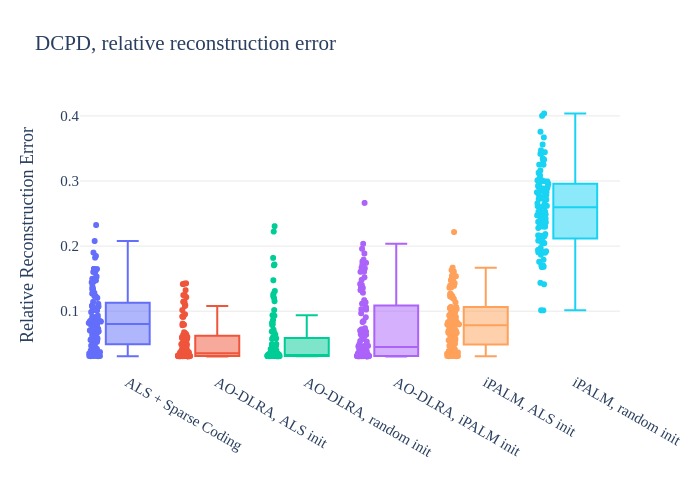}
	\includegraphics[width=0.49\textwidth]{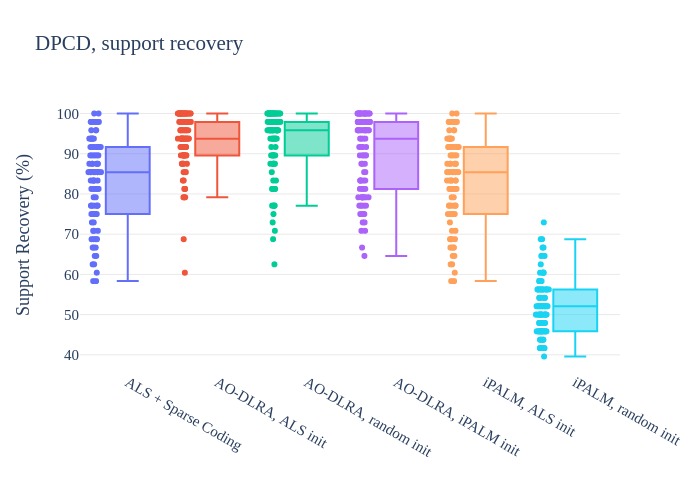}
  \caption{Relative reconstruction error (left) and support recovery (right, in \(\%\)) for DMF algorithms (top) and DCPD algorithms (bottom). One shared initialization is used for the randomly initialized methods, and 100 instances of each problem are tested. AO-DLRA, iPALM init is when the result of iPALM, random init is used as an initialization for AO-DLRA.}\label{fig:xp3}
\end{figure}

Figure~\ref{fig:xp3} shows the obtained results. It can be observed that both in DMF and DCPD, iPALM initialization is not significantly better than random initialization, and iPALM in fact provides quite poor results on its own. For DCPD, support recovery and reconstruction error is always better with AO-DLRA than when sparse coding the output of the ALS, even with random initialization. Overall, these experiments show that while the DLRA problem is challenging (the optimal support is almost always never found), reasonably good results are obtained using the proposed AO-DLRA algorithm. Furthermore, the support recovery scores of DMF using AO-DLRA are much higher than what could be obtained by chance, randomly picking elements in the support. This observation hints towards the identifiability of DMF. Remember indeed that without the dictionary constraint, matrix factorization is never unique when \( r>1 \), thus any posterior support identification for a matrix \( A \) given a factorization  \(Y=AB \) should fail.

\section{Conclusions and open questions}

In this manuscript, a Dictionary-based Low-Rank Approximation framework has been proposed. It allows to constrain any factor in a LRA to live in the union of k-dimensional subspaces generated by subsets of columns of a given dictionary. DLRA is shown to be useful for various signal processing tasks such as image completion or image denoising. A contribution of this work is an Alternating Optimization algorithm (AO-DLRA) to compute candidate solutions to DLRA. Additionally, the subproblem of estimating the sparse codes of a factor in a LRA, coined Mixed Sparse Coding, is extensively discussed. A heuristic convex relaxation adapted from LASSO is shown to perform very well for solving MSC when compared to other modified sparse coding strategies, and along the way, several theoretical results regarding MSC are provided.

There are several research directions that stem from this work. First, the identifiability properties of DLRA have not been addressed here. It was shown in previous works~\citep{Cohen2018Dictionary} that dictionary-based matrix factorization is identifiable when sparsity is exactly one, but the general case is harder to analyse. The identifiability analysis is furthermore model dependent while this work aims at tackling the computation of any DLRA. Second, while the proposed AO-DLRA algorithm proved efficient in practice, its convergence properties are lacking. It is reasonably easy to design an AO algorithm with guaranties for DLRA, but I could not obtain an algorithm with convergence guarantees which performance matched the proposed AO-DLRA. Finally, the proposed DLRA model could be extended to a supervised setting, where \( D \) is trained in a similar fashion to Dictionary Learning~\citep{Mairal2010Online}. This would mean computing a DLRA for several tensors with the same dictionary, a problem closely related to coupled matrix and tensor factorization with linearly coupled factors~\citep{schenker2020flexible}.

\section*{Conflict of Interest Statement}

The author declares that the research was conducted in the absence of any commercial or financial relationships that could be construed as a potential conflict of interest.

\section*{Author Contributions}
Jeremy E. Cohen is the sole contributor to this work. He studied the theory around Mixed Sparse Coding and Dictionary-based low-rank approximations, implemented the methods, conducted the experiments, and wrote the manuscript.

\section*{Funding}
This work is funded by ANR JCJC LoRAiA ANR-20-CE23-0010.

\section*{List of abbreviations}
The following abbreviations are used in the manuscript: Low-Rank Approximation (LRA), Canonical Polyadic Decomposition (CPD), Dictionary-based LRA (DLRA), Dictionary-based CPD (DCPD), Dictionary-based Nonnegative Matrix Factorization (DNMF), Nonnegative Matrix Factorization (NMF), Dictionary-based Matrix Factorization (DMF), Mixed Sparse Coding (MSC), Alternating Optimization (AO), Hard Thresholding (HT), Iterative Hard Thresholding (IHT), Orthognal Matching Pursuit (OMP), Fast Iterative Soft Thresholding Algorithm (FISTA), Hierarchical OMP (HOMP), Signal to Noise Ratio (SNR), Spectral Angular Mapper (SAM).

\section*{Acknowledgments}
The author thanks Rémi Gribonval for commenting an early version of this work.

\section*{Data Availability Statement}
The scripts used to generate the synthetic dataset for this study can be found alongside the code in the online repositories \url{https://github.com/cohenjer/mscode} and \url{https://github.com/cohenjer/dlra}.

\bibliography{all_refs}

\appendix

\section{Proofs}

\subsection{Proof of Proposition 1}
Recall the following problem:
\begin{equation}\label{eq:MSC2bisbis}
  \underset{\|Y-DXB^T\|^2_F\leq \epsilon_1}{\min} \max_i\|X_i\|_0~.
\end{equation}

Let \( X^\ast \in\mathbb{R}^{d\times r} \) the unique solution to~\eqref{eq:MSC2bisbis} and set \(k = \max_{i\leq r} \|X^\ast_i \|_0\). Let \( Z\in\mathbb{R}^{d\times r} \) such that for any \( i\leq r \), \( \|Z_i\|_0\leq k \) and \( Z\neq X^\ast \). If
\begin{equation}
  \|Y - DZB^T \|_F^2 \leq \| \|Y - DX^\ast B^T\|_F^2
\end{equation}
then \( Z \) is an admissible solution for~\eqref{eq:MSC2bisbis} with lower cost that \( X^\ast \) which contradicts its uniqueness. Thus, any admissible solution \( Z \) different from \( X^\ast \) has strictly larger residual, and \( X^\ast \) is the unique solution to MSC. The same proof holds conversely by swapping the costs and constraints.
\vspace{1em}

\subsection{Proof of Proposition 2}
First notice that the spark condition on \( D \) implies uniqueness of the solution \( X \) for a fixed support \( \Supp(X) \). Indeed, if there exist two different solutions \(X,X'\) with the same support, then \( DXB^T = DX'B^T \) implies \( D\left(X-X'\right) = 0 \) since \( B \) is full column rank. In turn this means that \( X - X' \) is a columnwise \(2k\)-sparse matrix, whose columns are in the Kernel of a spark \(2k\) dictionary. Therefore we must have \( X=X' \).

Furthermore, the sets \( \Omega_S = \{DXB^T,~ \Supp(X)=S \} \) for \( |S|\leq k \) have trivial intersection. Indeed, if \( Y \) is a matrix in two different \( \Omega_S \), then by the same chain of arguments, \( Y = DXB^T = DX'B^T \) yields \( X = X' \) which contradicts that the supports should be different.

Therefore, MSC can be recast as
\begin{equation}
  \text{Project } Y \text{ on the union of sets } \cup_{|S|\leq k} \Omega_S
\end{equation}

One may notice that the sets \( \Omega_S \) are strict subspaces as long as \( k<d \). For some matrix \( Y \) to admit two closest sets \( \Omega_S \), it must live in the set of median planes of pairs of \( \Omega_S \), which are also strict subspaces. There is only a finite number of such median planes as well, and therefore the set of \( Y \) that are equidistant to several \( \Omega_S \) sets has zero Lebesgue measure.

To summarize, most \(Y\) have a single closest \( \Omega_S \), and their representation \( X \) is unique because of the spark hypothesis.
\vspace{1em}

\subsection{Proof of Lemma 1}
We start by upper-bounding
\( \|DX - DX' \|^2_F = \|DX - YB\left(B^TB\right)^{-1} + YB\left(B^TB\right)^{-1} - DX' \|_F^2  \leq \|YB\left(B^TB\right)^{-1} - DX \|_F^2 + \delta \).
Using the fact that \( \|Y - DXB \|^2_F \leq \epsilon  \) implies \( \|YB\left(B^TB\right)^{-1} - DX \|_F^2 \leq \sigma^2_{\max}(B\left(B^TB\right)^{-1})\epsilon = \sigma^2_{\min}(B)\epsilon \), it therefore holds that
\( \|DX - DX' \|^2_F \leq \sigma_{\min}(B)\epsilon + \delta \).
Because \( X \) and \( X' \) are both columnwise \(k\)-sparse, \( X - X' \) belongs to a subspace of dimension at most \( 2k \) and at worse,
\( \|D (X - X') \|_F^2 \geq (\sigma_{\min}^{(2k)}(D))^2 \|X - X' \|_F^2 \). Rearranging terms yields the proposed bound.
\vspace{1em}

\subsection{Proof of Proposition 3}
Suppose that \( X \) and \( X' \) have different supports, then there exist at least one column and two entries with distinct row indices \( X_{ij} \) and \( X'_{{i'}j} \) such that \( X_{{i'}j}= X'_{ij} = 0 \).  This means that \( \|X - X' \|_F^2 \geq {X_{ij}}^2 + {X'_{{i'}j}}^2\). Therefore, under the hypotheses of Lemma 1, it must hold that
\begin{equation}
 \sqrt{{X_{ij}}^2 + {X'_{{i'}j}}^2} \leq \frac{1}{\sigma^{(2k)}_{\min}(D)}\sqrt{\delta+\frac{\epsilon}{\sigma^2_{\min}(B)}}
\end{equation}
By contraposition, if the above inequality is reversed for any collection of pairs of indices \( \{(i, {i'}\}_{j\leq r} \) then \( X \) and \( X' \) have the same support. Using the worse case as a lower bound yields
\begin{equation}
      \min_{j\leq r}\sqrt{\min_{i\in \Supp(X_j)} {X_{ij}}^2 + \min_{ i'\in \Supp(X_j')} {X'_{i'j}}^2} >  \frac{1}{\sigma^{(2k)}_{\min}(D)}\sqrt{\delta+\frac{\epsilon}{\sigma^2_{\min}(B)}}.
\end{equation}
\vspace{1em}

\subsection{Proof of Proposition 4}
Suppose that there exist a column indexed by\( i\leq r \) such that \( D_{S_i} \) is not full column rank. Then any \( Z = X^\ast + t[0,\ldots, 0, \overline{c},0,\ldots 0] \) with \( c\in \text{Ker}(D_{S_i}) \), \( \overline{c} \) an embedding of \( c \) in the ambiant space positioned in the $i$-th position and \( t\in\mathbb{R} \) has the same residuals \( \|Y - DZB^T\|_F^2 \) than \( X^\ast \). Moreover, the map \( t\mapsto \| Z_i \|_1 \) is piece-wise constant without a breakpoint at \( t=0 \). Therefore there exist a \( t \) small enough such that \( \|Z_i\|_1 \leq \|X^\ast\|_1 \) which contradicts the fact that \( X^\ast \) is a unique solution.

This proof is similar to the proof of Proposition 7, which is more detailled.
\vspace{1em}

\subsection{A useful lemma}

\begin{lemma}
  We prove a general result studying
  \begin{equation}
    \phi(X) = \|Y - DXB^T \|_F^2 + g(X)
  \end{equation}
  for any convex map \( g \). Let us show that \( DXB^T=DZB^T \) and \( g(X) = g(Z) \) for any two minimizers \( X,Z \).

  Let \( X \) and \( Z \) two minimizers of the above cost such that \( \phi(X)=\phi(Z)=c \) and suppose by contradiction that \( DXB^T \neq DZB^T \). Then because \( \psi: X\mapsto \|Y-X \|_2^2 \) is strongly convex, for any \( t\in(0,1) \),
  \begin{equation}
  \begin{aligned}
    \phi(tX + (1-t)Z) &=  \psi(t DXB^T + (1-t)DZB^T) + g(t X + (1-t)Z) \\
    &< t\left[\psi(DXB^T)+ g(X)\right] + (1-t)\left[\psi(DZB^T) + g(Z)\right] \\ &< c
  \end{aligned}
\end{equation}
and we have just shown that \( tX + (1-t)Z \) reaches a lower cost than the minimal cost \( c \) which is absurd. The fact that \( g \) is also constant for all minimisers is a direct consequence.
\end{lemma}
\vspace{1em}

\subsection{Proof of Proposition 5}
First, suppose that \( X^{\ast}=0 \) is a solution to the Block Lasso problem. Then it is a fixed point of the proximal iteration
\begin{equation}
  X^{\ast} = \text{S}_{\lambda}( X^{\ast} + D^TYB - D^TDX^{\ast}B^TB)
\end{equation}
which yields in that case
\begin{equation}
  \forall i\leq r, \;\text{S}_{\lambda_i}(D^TYB_i) = 0
\end{equation}
This imposes that \( \lambda_i \geq \lambda_{i,\max} := \|D^TYB_i\|_{\infty} \).
Using the above lemma, we can conclude that any other solution \( Z \) than zero must satisfy \( \|Z_i\|_{1}=0 \) for all columns \( i\leq r \) which immediately implies that \( Z=0 \).
\subsection{Proof of Proposition 6}
Let \( f(X) = \max_i \|X_i\|_0 \) defined on \( [-1,1]^{d\times r} \). We are going to compute the double Fenchel conjugate \( f^{\ast\ast} \) where by definition for any real map \( h \), the convex conjugate is defined as \( h^{\ast}(Y) = \sup_{X\in[-1,1]^{d\times r}} \langle Y, X \rangle - h(X)\).

We have
\begin{equation}\label{eq:fc}
      f^{\ast}(Y) = \sup_X \langle Y,X\rangle - \max_i \|X_i\|_0.
\end{equation}
Let us suppose that \( k=\max_i \|X_i\|_0 \) is fixed. Then we solve a slightly different problem
\begin{equation}
  \sup_{\|X_i\|_0\leq k, X\in[-1,1]^{d\times r}} \langle Y,X \rangle
\end{equation}
which is easily shown to be solved by \( X^{\ast} \) being equal to \( \text{sign}(Y) \) for the \( k \) largest absolute values in \( Y \) and \( 0 \) otherwise. In other words,
\begin{equation}
  \sup_{\|X_i\|_0\leq k, X\in[-1,1]^{d\times r}} \langle Y,X \rangle =
  \|H_k(Y)\|_1
\end{equation}
where \( H_k \) is the column-wise hard-thresholding operator. Furthermore, splitting the minimization in~\eqref{eq:fc} over \( k=\max_i \| X_i\|_0 \) and \( X \) with fixed \( k \), it holds that
\begin{equation}\label{eq:fc_int}
  f^{\ast} = \max_{k\in[0,d]} \|H_k(Y)\|_1 - k
\end{equation}
Sweeping over the various values of \( k \), one can notice that adding one to \( k \) means adding \( \sum_i|Y_{\sigma_i(k+1)i}|\) and substracting one to the cost in~\eqref{eq:fc_int}, where \( \sigma_i \) sorts decreasingly the \( i\)-th column of \( Y \). Therefore, as long as \( \sum_i |Y_{\sigma_{i}(k)i}| \geq 1 \), increasing \( k \) increases the cost, while the converse condition decreases the cost. We can therefore conclude that
\begin{equation}
  f^{\ast}(Y) = \| H_{k(Y)}(Y)\|_1 - k(Y), \;\; k(Y)=\max \{k ~|~ \sum_{i=1}^{r} |Y_{\sigma_i(k)i}| \geq 1 \}
\end{equation}
which may be equivalently rewritten as
\begin{equation}
  f^{\ast}(Y) = \sum_{j=1}^{d} \mathds{1}_{\sum_i |Y_{\sigma_i(j)i}| \geq 1} \left[ -1 + \sum_{i=1}^{r} |Y_{\sigma_i(j)i}| \right]
\end{equation}

Now for any \( X\in[-1,1]^{d\times r} \), let us compute
\begin{align}
  f^{\ast\ast}(X) =& \sup_{Y}~ \langle Y,X \rangle - f^{\ast}(Y) \\
                  =& \sup_{Y}~ \sum_{ji} Y_{\sigma_i(j)i}X_{\sigma_i(j)i} -
                  \sum_{j=1}^{d} \mathds{1}_{\sum_i |Y_{\sigma_i(j)i}| \geq 1} \left[ -1 + \sum_{i=1}^{r} |Y_{\sigma_i(j)i}| \right]
\end{align}
After some manipulations, we obtain the following equivalent problem:
\begin{equation}
  \sup_{Y} \sum_{j=1}^{d}\sum_{i=1}^{r} \left[ X_{\sigma_i(j)i} - \mathds{1}_{\sum_i |Y_{\sigma_i(j)i}| \geq 1}\text{sign}(Y_{\sigma_i(j)i}) \right] Y_{\sigma_i(j)i} +  \sum_{j=1}^{d} \mathds{1}_{\sum_i |Y_{\sigma_i(j)i}| \geq 1},
\end{equation}
which is separable in \( j \). Therefore, one only needs to maximize for each \( j \) the summand
\begin{equation}
  \sum_{i=1}^{r} \left[ X_{\sigma_i(j)i} - \mathds{1}_{\sum_i |Y_{\sigma_i(j)i}| \geq 1}\text{sign}(Y_{\sigma_i(j)i}) \right] Y_{\sigma_i(j)i} +  \mathds{1}_{\sum_i |Y_{\sigma_i(j)i}| \geq 1}.
\end{equation}
Let us first suppose that \( \sum_i |Y_{\sigma_i(j)i}| \geq 1 \). Then we can see that the coefficient before \( Y_{\sigma_i(j)i} \) is always of opposite sign to \( Y_{\sigma_i(j)i} \) and therefore the cost always decreases with \( |Y_{\sigma_i(j)i}| \) for all \( i \). Secondly, supposing that \( \sum_i |Y_{\sigma_i(j)i}| \leq 1 \), the cost is simply a scalar product between \( X \) and \( Y \) which can be maximized by setting \( Y \) proportional to the sign of \( X \) and increasing \( |Y| \).

Therefore, we can conclude that
\begin{equation}
  f^{\ast\ast}(X) = \sup_{\forall j,\;\sum_{i=1}^{r}|Y_{\sigma_i(j)i}| =1 } \langle Y,X\rangle
\end{equation}
At this point, one may notice that the set \( \{Y~|~ \forall j,\;\sum_{i=1}^{r}|Y_{\sigma_i(j)i}| =1\} \) is very particular. Indeed, since \( \sum_i |Y_{\sigma_i(1)i}| = 1 \) and for all \( i,j \), \( Y_{\sigma_i(j)i} \leq Y_{\sigma_{i}(1)i} \), we must have
\begin{equation}
  |Y_{\sigma_i(j)i}| = |Y_{\sigma_i(1)i}|
\end{equation}
and therefore the columns of \( Y \) in that set are constant up to sign.

Finally, the double conjugate yields
\begin{equation}
  \sup_{\sum_i y_i = 1,\; y_i\geq 0,\; \epsilon_{ji}\in\{-1,1\}} \sum_i y_i  \left( \sum_j \epsilon_{ji} X_{ji} \right)
\end{equation}
which supremum can be easily shown to be attained for \( \epsilon_{ji_\text{max}} = \text{sign}(X_{ji_\text{max}}) \), \( y_{i_\text{max}} = 1 \) and \( y_{i\neq i_\text{max}} = 0 \), where \( i_\text{max} \) is the index of the column of \( X \) with largest \( \ell_1 \) norm. We can thus finally conclude that
\begin{equation}
  f^{\ast\ast}(X) = \max_i \|X_i\|_1 = \|X\|_{1,1}
\end{equation}
\vspace{1em}


\subsection{Proof of Proposition 7}

Assume by contradiction that for any \( i \) in \( \mathcal{I} \), there exist a vector \( \tilde{c}_i\in\text{Ker}D_{S_i} \), so that \( D_{S_i}\tilde{c}_i = 0 \). Define \( c_i \) as \( \tilde{c}_i \) on \( S_i \) and \(0\) on \( \overline{S}_i \), and the matrix \( C \) of concatenated \( c_i \) vectors. By construction, \( DC = 0 \) and thus for any \( t\in\mathbb{R} \),
 it holds that \( D(X^{\ast} + tC)B^T = DX^{\ast}B^T \).

Now let us check that we can build some small \( t \) such that \( \| X^{\ast}+tC\|_{1,1} \leq \| X^{\ast}\|_{1,1} \) which will contradict the uniqueness of \( X^{\ast} \).

Let us fix some index \( i\in\mathcal{I} \). It can be observed that the map \( \phi_i: t\mapsto \|X^{\ast}_i + tc_i \|_1 \) is continuous and piece-wise affine. For any index \( j \) outside the support \( S_i \), both \( X^{\ast}_{ij} \) and \( c_{ij} \) are zero. Thus the slope of \( \phi_i \) changes when there exist \( j\in S_i \) and \( t\in\mathbb{R}\) such that \( X^{\ast}_{ij}+ t c_{ij} = 0 \). Therefore, the slope does not change at \( t=0 \).

Now consider the slope of each map \( \phi_i \) around \(0\). If that slope is negative for some index \( i \), we can always replace \( c_i \) with \( -c_i \) the slope around \(0\) becomes positive. Therefore, one may assume that \( t\mapsto \| X^{\ast} + tC \|_{1,1} \) has positive slope around zero, and for a small \( t<0 \),
\begin{equation}
  \|X^{\ast}+tC \|_{1,1} \leq \|X^{\ast} \|_{1,1}~.
\end{equation}

Now suppose that \( D \) is overcomplete. Suppose by contradiction that there exist \( i\leq r \) such that \( \|X_i^\ast \|_1 < \|X^\ast\|_{1,1} \). Then with similar notations, any \( Z = X^\ast + tC \) with \( t \) small enough has the same cost as \( X^\ast \), which is impossible since \( X^\ast \) is the unique solution. Therefore \( \mathcal{I}=[1,\ldots,r] \).
\vspace{1em}


%
\subsection{Proof of Proposition 8}
First, suppose that \( X^{\ast}=0 \) is a solution to the Mixed Lasso problem. Then it is a fixed point of the proximal iteration
\begin{equation}
  X^{\ast} = \prox_{\lambda,\ell_{1,1}}( X^{\ast} + D^TYB - D^TDX^{\ast}B^TB)
\end{equation}
which yields in that case
\begin{equation}
  \prox_{\lambda,\ell_{1,1}}(D^TYB) = 0
\end{equation}
According to Proposition 5 in~\cite{cohen2020computing}, this imposes that \( \lambda \geq \sum_{i=1}^{r} \|D^TYB_i \|_{\infty} \).
At this stage, we may use the above Lemma to conclude that any other solution \( Z \) than zero must satisfy \( \|Z\|_{1,1}=0 \) which immediately implies that \( Z=0 \).
\vspace{1em}


\section{Additional experiments for Mixed Sparse Coding}

Unless specified otherwise, these additional experiments use the same settings as the experiments on Mixed Sparse Coding.
\vspace{1em}

\subsection{Additional test 1: Solving nonnegative MSC with Block-FISTA and nonnegative Block-FISTA}
\begin{figure}
  \centering
  \includegraphics[width=\textwidth]{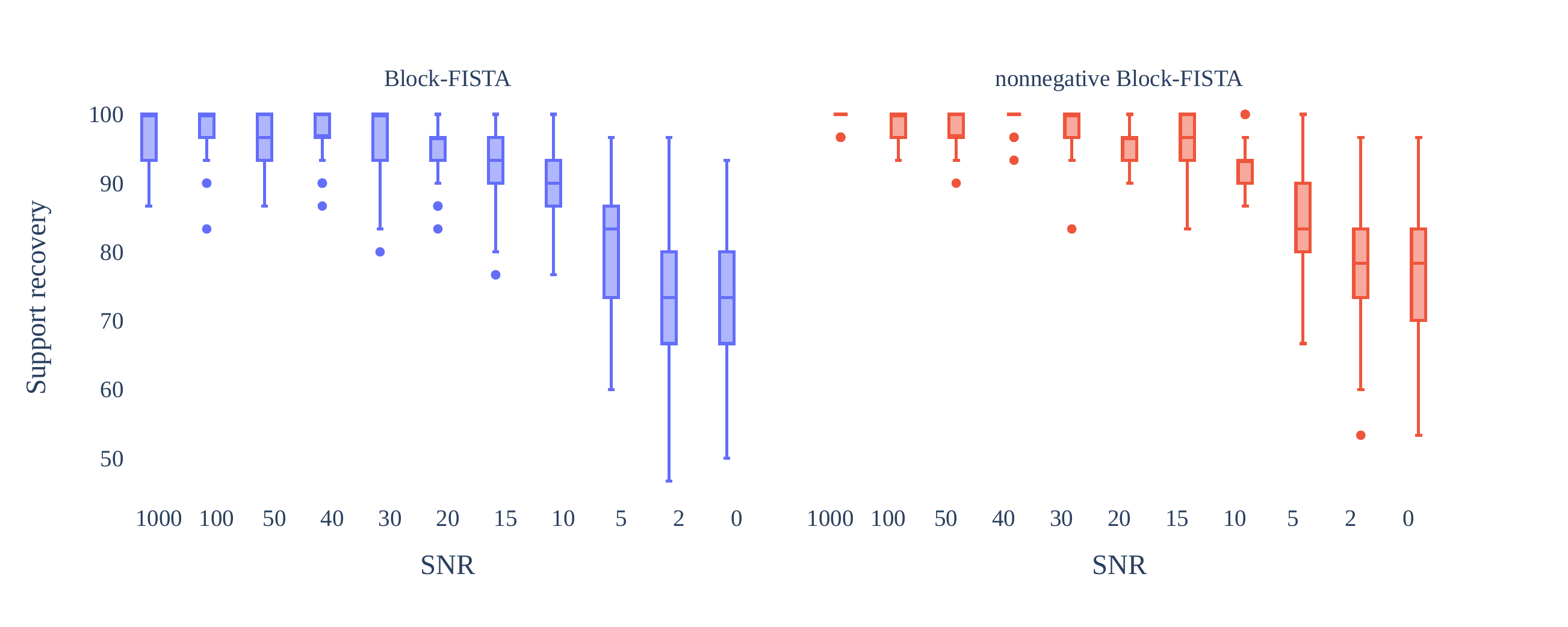}
  \caption{Support recovery against noise, for Block-FISTA and nonnegative Block-FISTA.}
  \label{fig:Testnn}
\end{figure}

Let us compare the performance of Block-FISTA and its nonnegative variant on the scenario of Test 1, with the nonzero elements in \( X \) sampled from a Uniform distribution on \([0,1]\). Indeed, in the presence of noise, it is likely that estimated coefficients \( X \) without nonnegativity constraint will take undesired negative values, and therefore the nonnegative setting is expected to perform slightly better in terms of support recovery. Only Block-FISTA and nonnegative Block-FISTA  are compared since the other algorithms for solving MSC are already studied in other tests. Contrarily to Test 1, the same regularization amount is used for Block-FISTA and the nonnegative variant, since the two methods are very similar and they can thus be compared  in similar settings. Figure~\ref{fig:Testnn} shows that the nonnegative Block-FISTA is slightly outperforming its unconstrained counterpart, especially for noisy scenarios.
\vspace{1em}

\subsection{Additional test 2: support recovery vs the conditioning of B}
\begin{figure}
  \includegraphics[width=\textwidth]{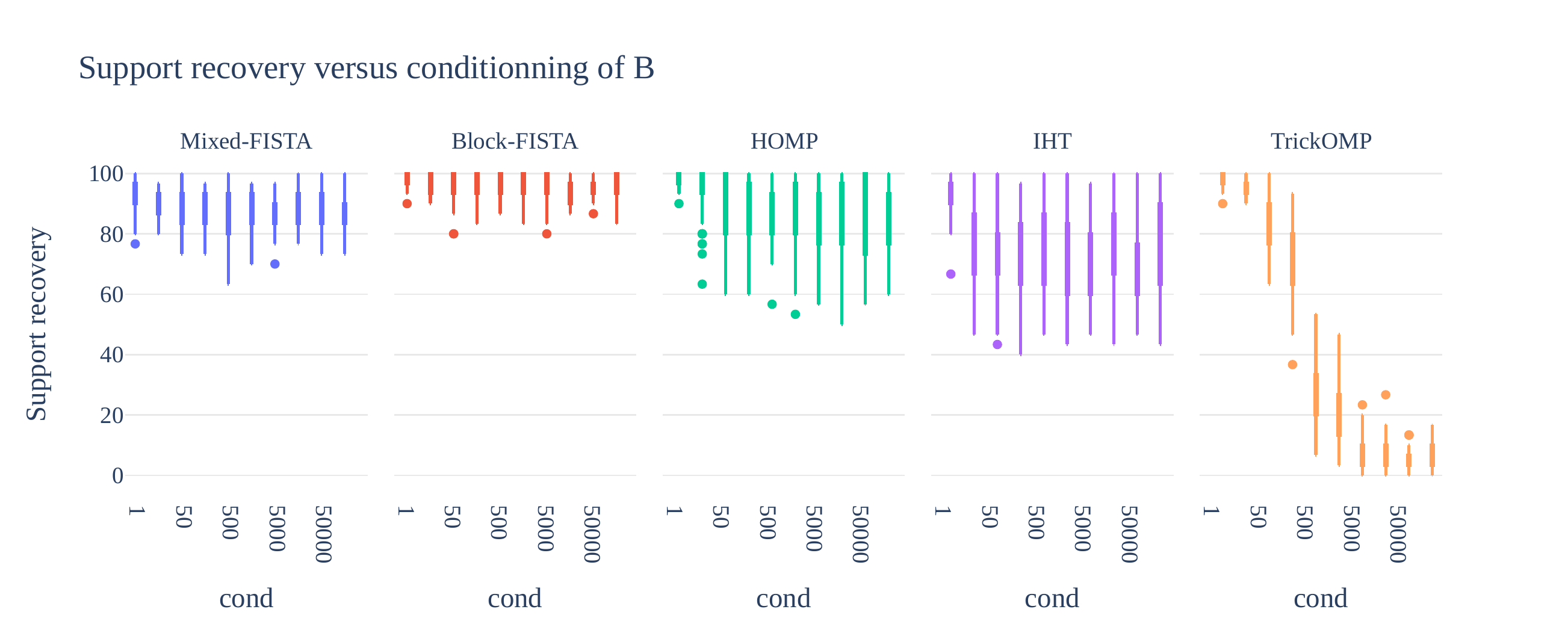}
  \caption{Support recovery (in \( \% \)) of the proposed heuristics to solve MSC at various conditionings for matrix \( B \). A total of \( 50 \) problems instances are used, with a single random but shared initialization for all methods.}
  \label{fig:add2}
\end{figure}

In this experiment, the conditioning of mixing matrix \( B \), which plays a role in the precision of the method TrickOMP according to Proposition~3, varies on the grid \( [1, 10, 50, 100, 5\times10^{2}, 10^3, 5\times10^3, 10^4, 5\times10^4, 10^5] \). Results are shown in Figure~\ref{fig:add2}.

As expected TrickOMP does not behave well when factor matrix \( B \) is ill-conditioned. HOMP and IHT have interestingly nice performance when the conditioning is exactly 1, but weaker performance as soon as the conditioning gets worse. On the other hand, FISTA methods are not visibly affected by the conditioning of \( B \).
\vspace{1em}

\subsection{Additional test 3: support recovery with various initializations}
\begin{figure}
  \centering
  \includegraphics[width=\textwidth]{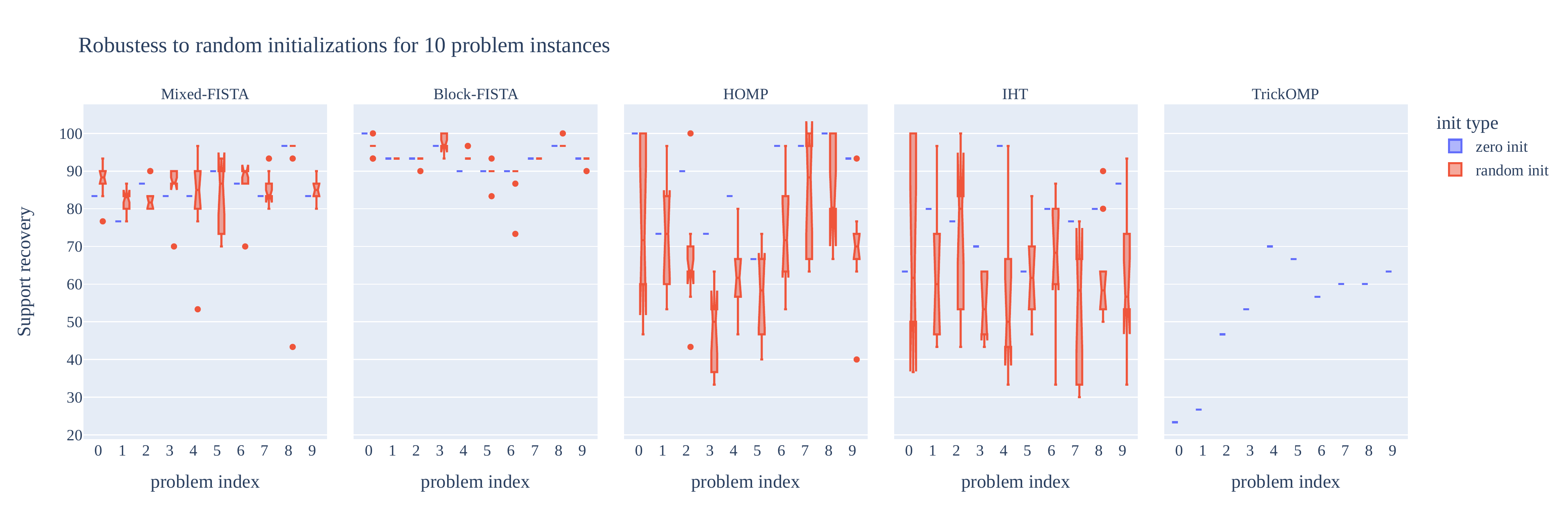}
  \caption{Support recovery (in \( \% \)) of the proposed heuristics to solve MSC with 10 randomly generated problems, with 10 random initializations each (red boxes). The zero initialization is also tested (dashes in blue)}
  \label{fig:Test4}
\end{figure}

In this test, both the dependence on initialization for all algorithms as well as variation of performance across \( M=10 \) problem instances are studied. All algorithms use the same initial values. The initial entries estimates \( X \) are generated using either standard Gaussian distribution or only zeros. For the Gaussian test, I tried \(N = 10\) initializations. Figure~\ref{fig:Test4} shows the results for these three initialization methods.

From these results, it can be observed that for all methods, choosing an initial \( X \) as a zero matrix yields generally good results although there can be some benefit to random initializations. Moreover as expected, the convex regularization-based methods are not much dependent on initialization. Overall, as observed also in the other tests, Block-FISTA is performs more consistently than for instance HOMP which may fail at solving some problem instances. A noticeable property of IHT is that they are heavily dependent on initialization, and while they may perform very well if a good starting point is provided, most of the runs however fail when compared to Block-FISTA.
\vspace{1em}

\subsection{Additional test 4: support recovery with regularization level for convex methods}
It is important to provide some concrete idea of the sensitivity of the performance of Block-FISTA and Mixed-FISTA to the choice of the regularization amount \( \alpha \). Recall that for the Block-FISTA, although in principle there are \( r \) values \( \lambda_i \) to choose, in all the tests I use a single ratio \( \lambda_i = \alpha\lambda_{i,max} \). For the Mixed-FISTA, again recall that the regularization parameter is given by a ratio \( \alpha\lambda_{max} \) and therefore for both methods, only the regularization amount \(\alpha\) is tuned. Below, the value of \( \alpha\in[0,1] \) always refers to this ratio.

\begin{figure}
  \centering
  \includegraphics[width=\textwidth]{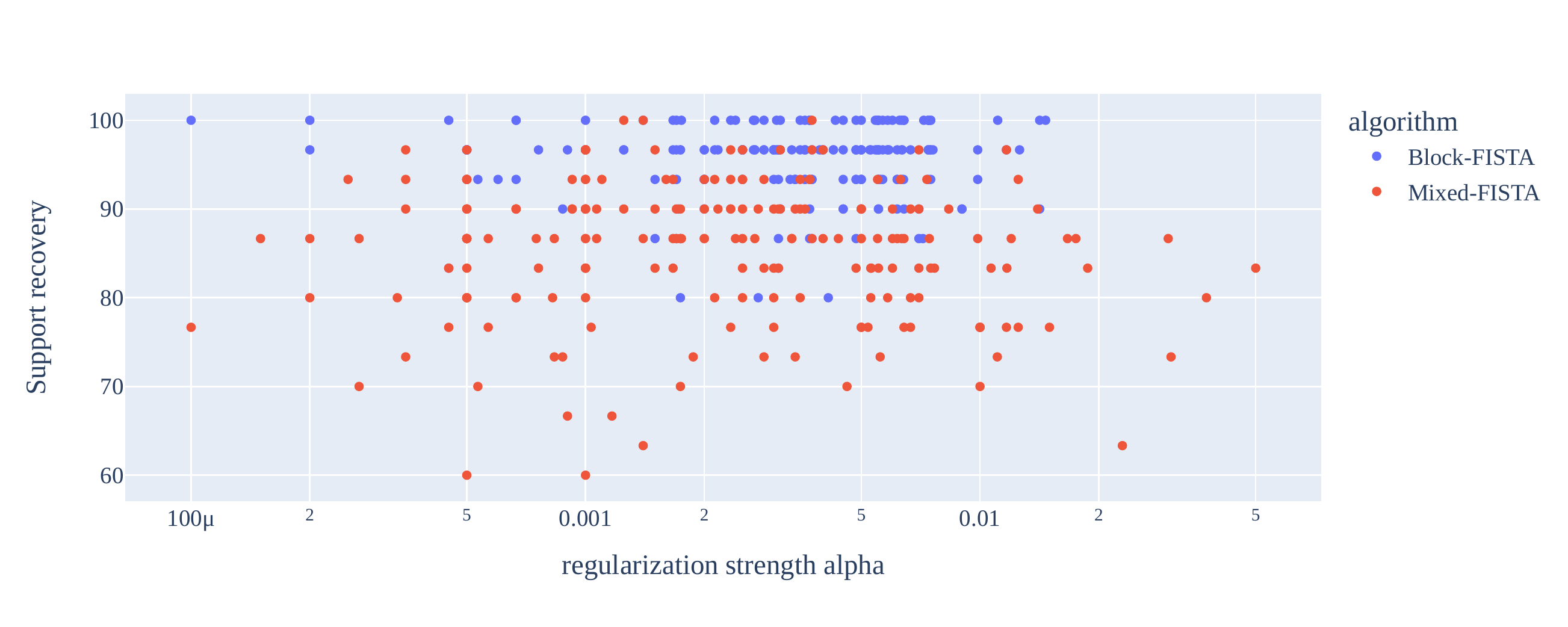}
  \caption{Performance of the convex relaxation algorithms in terms of support recovery (in \%). The heuristic optimal regularization percentage \( \alpha \) is given by the x axis. Each dot is the score for one instance of the mixed sparse coding problem solved with Block-FISTA (blue) or Mixed-FISTA (red).}
  \label{fig:Test5-1}
\end{figure}

\begin{figure}
  \includegraphics[width=\textwidth]{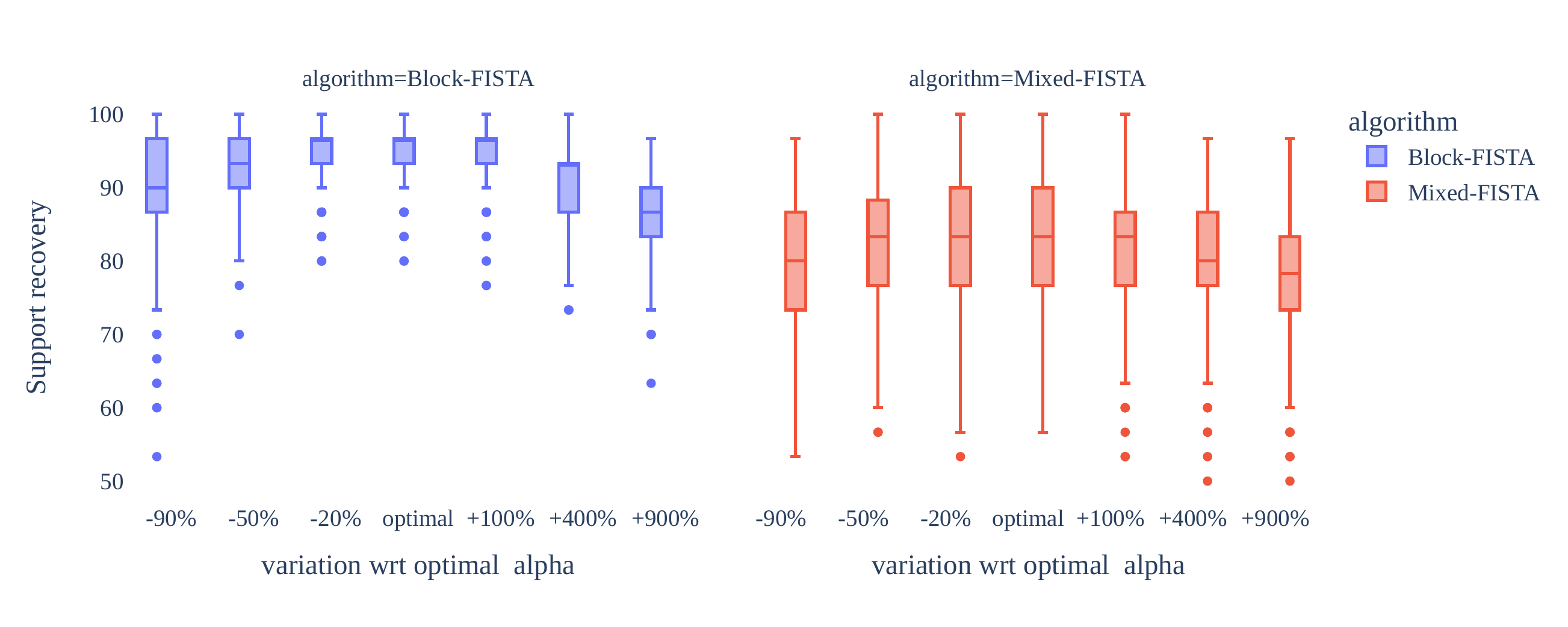}
  \caption{Performance of Block-FISTA and Mixed-FISTA when the regularization amount \( \alpha \) is chosen around the optimal value.}
  \label{fig:Test5-2}
\end{figure}

Two questions come to mind when studying the sensitivity to \( \alpha \): are the best values of \(\alpha\) changing significantly given a fixed experimental setup (problem sizes essentially), and how are performance degraded when a suboptimal \( \alpha \) is chosen? To answer these two questions, first the best \( \alpha \) values in terms of support recovery are computed on \( 200 \) runs for both algorithms, with \( \alpha \) on a grid \( [0, 0.0001, 0.0002, 0.0005, 0.001, 0.002, 0.005, 0.01, 0.02, 0.05, 0.1] \). Then for each simulation, both algorithms are again ran with regularization on a grid \[ [-90\%,-50\%,-20\%,'optimal',+100\%,+400\%,+900\%]\] relative to the best \( \alpha \) found in the previous step. By \( +n\% \), it is meant that \(\alpha = \alpha_{\text{optimal}}(1+n/100)\).
Figure~\ref{fig:Test5-1} reports the best \( \alpha \) found for the \( N \) runs, while Figure~\ref{fig:Test5-2} shows the results when deviating from these best \( \alpha \)s.

It can be observed that while the oracle best value of \( \alpha \) can change quite significantly for several similarly generated problems, the value of \( \alpha \) does not need to be optimal for performances to be optimal, in particular with Block-FISTA.

\section{Additional Pseudo codes}
\begin{algorithm}
  \caption{accelerated IHT for Mixed Sparse Coding}\label{alg:iht}
  \begin{algorithmic}
   \STATE{\textbf{Input:} data \(Y\), dictionary \(D\), mixing matrix \(B\), sparsity level \(k\), initial value \(X\).}
   \STATE{\textbf{Output:} estimated codes \( X \), support \( S \).}
   \STATE{Precompute \(D^TD, D^TYB\) and \( B^TB \) if memory allows.}
   \STATE{Compute stepsize \( \eta = \frac{1}{\sigma(D)^{2}\sigma(B)^{2}} \)}
   \STATE{Initialize \(Z=X, \; \beta=1\).}
   \WHILE{stopping criterion is not reached}
    \STATE{\(X_{\text{old}}=X\)}
    \STATE{\(X = \text{HT}_{k}\left( Z - \eta(D^TDZB^TB - D^TYB) \right)\)}
    \STATE{\(\beta_{\text{old}} = \beta\)}
    \STATE{\( \beta = \frac{1}{2}(1+\sqrt{1+4\beta^2})\)}
    \STATE{\( Z = X + \frac{\beta_{\text{old}}-1}{\beta} \left( X-X_{\text{old}} \right) \)}
   \ENDWHILE
   \STATE{Estimate the support \( S=S(X) \)}
   \STATE{Set \( X \) as the least squares solution
   with support \( S \).}
  \end{algorithmic}
\end{algorithm}

\begin{algorithm}
  \caption{FISTA for Mixed Lasso}\label{alg:fista}
  \begin{algorithmic}
   \STATE{\textbf{Input:} data \(Y\), dictionary \(D\), mixing matrix \(B\), regularization ratio \(\alpha\in[0,1]\), sparsity level \(k\), initial value \(X\).}
   \STATE{\textbf{Output:} estimated codes \( X \), support \( S \).}
   \STATE{Precompute \(D^TD, D^TYB\) and \( B^TB \) if memory allows.}
   \STATE{Compute \( \lambda_{\max} \) as in Proposition 8, 
   and set \( \lambda = \alpha\lambda_{\max}\)}
   \STATE{Compute stepsize \( \eta = \frac{1}{\sigma(D)^{2}\sigma(B)^{2}} \)}
   \STATE{Initialize \(Z=X, \; \beta=1\).}
   \WHILE{stopping criterion is not reached}
    \STATE{\(X_{\text{old}}=X\)}
    \STATE{\(X = \text{prox}_{\eta\lambda\ell_{1,1}}\left( Z - \eta(D^TDZB^TB - D^TYB) \right)\)}
    \STATE{\(\beta_{\text{old}} = \beta\)}
    \STATE{\( \beta = \frac{1}{2}(1+\sqrt{1+4\beta^2})\)}
    \STATE{\( Z = X + \frac{\beta_{\text{old}}-1}{\beta} \left( X-X_{\text{old}} \right) \)}
   \ENDWHILE
   \STATE{Estimate the support \( S=S(X) \)}
   \STATE{Set \( X \) as the least squares solution
   with support \( S \).}
  \end{algorithmic}
\end{algorithm}

\begin{algorithm}
  \caption{An inertial PALM algorithm for DLRA (iPALM)}\label{alg:ipalm}
  \begin{algorithmic}
   \STATE{\textbf{Input:} Initial guesses \( X^{(0)}, B^{(0)} \), data \( Y \), dictionary \( D \), sparsity level \( k\leq n \), iteration number \( l_{\max} \), stepsize safegard \( \mu \leq 1 \).}
   \STATE{\textbf{Output:} Final estimated factors \( X^{(l)} \) and \( B^{(l)} \)}
   \STATE{Precompute \( D^TD, \;D^TY \) and \( \epsilon_{D} = \|D^TD\|_F \) if memory allows.}
   \STATE{Set \(l=0\), initialize \( Z^{(0)}=X^{(0)} \).}
   \WHILE{ stopping criterion is not reached }
     \STATE{\(l=l+1\)}
     \STATE{\underline{\( B \) update: }}
     \STATE{Update \( B^{(l)}\in\Omega_B \) using block-wise gradient updates.}
     \STATE{\underline{\(X\) update:}}
     \STATE{Old estimate storage for inertia: \(X_{\text{old}} = X^{(l-1)}\)}
     \STATE{Acceleration heuristic from iPALM: \( \beta = \frac{l-1}{l+2} \)}
     \STATE{Stepsize evaluation: \( \eta^{(l)}= \frac{\mu}{\epsilon_{D}\|{B^{(l)}}^TB^{(l)}\|_F} \).}
     \STATE{Compute efficiently data-factor product\( (D^TY)B^{(l)} \) and inner products \( {B^{(l)}}^TB^{(l)} \).}
     \STATE{\(X^{(l)} = \text{HT}_{k}\left(Z^{(l-1)} - \eta^{(l)}\left(D^TDZ^{(l-1)}{B^{(l)}}^TB^{(l)} - D^TYB^{(l)} \right) \right)\)}
     \STATE{\( Z^{(l)} = X^{(l)} + \beta\left(X^{(l)} - X_{\text{old}}\right) \)}
  \ENDWHILE
  \end{algorithmic}
\end{algorithm}

%

%

\end{document}